\documentclass[runningheads]{llncs}

\usepackage{eccv}

\usepackage{eccvabbrv}

\usepackage{graphicx}
\usepackage{booktabs}
\usepackage{amsmath}
\usepackage{listings}
\usepackage{tcolorbox}
\tcbuselibrary{listings}

\DeclareMathOperator*{\argmin}{argmin}

\usepackage[accsupp]{axessibility}  

\usepackage{hyperref}

\usepackage{orcidlink}

\begin{document}

\title{3DreamBooth: High-Fidelity 3D Subject-Driven Video Generation Model}





\author{Hyun-kyu Ko\inst{1}$^\ast$ \and
Jihyeon Park\inst{2}$^\ast$ \and
Younghyun Kim\inst{1} \and \\
Dongheok Park\inst{2} \and
Eunbyung Park\inst{1}$^\dagger$}

\authorrunning{Ko et al.}

\institute{$^1$Yonsei University \quad $^2$Sungkyunkwan University
\url{https://ko-lani.github.io/3DreamBooth}}

\maketitle
\let\thefootnote\relax\footnotetext{$^\ast$Equal contribution}
\let\thefootnote\relax\footnotetext{$^\dagger$Corresponding author}

\begin{figure}[ht]
\centering
\includegraphics[width=0.99\textwidth]{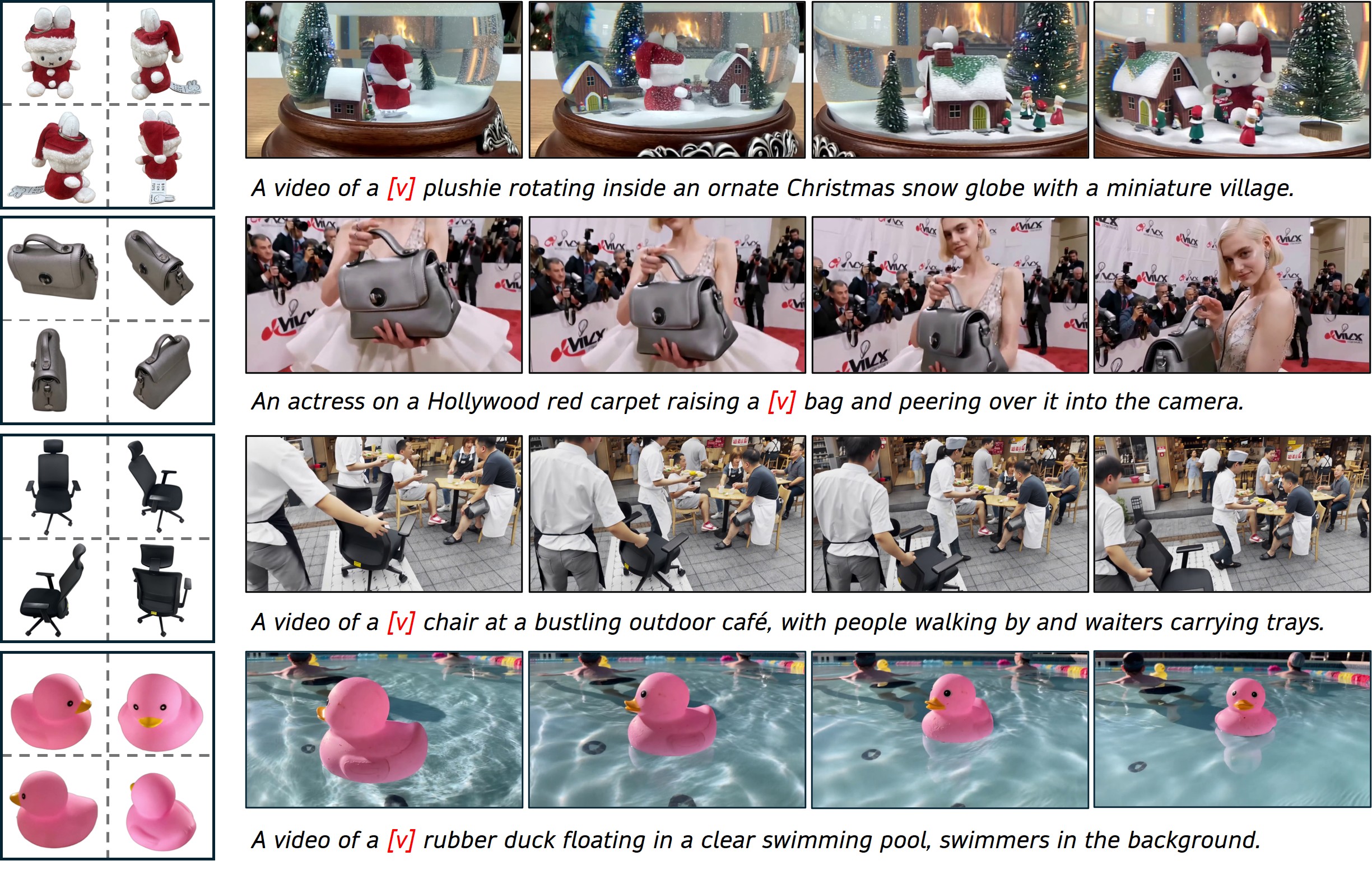}
\caption{\textbf{3D-aware video customization using our proposed framework.} Given a few multi-view reference images (left) and a text prompt, our approach generates high-fidelity, view-consistent videos that seamlessly integrate customized 3D subjects into dynamic environments.}
\label{fig:demo}
\end{figure}

\begin{abstract}
Creating dynamic, view-consistent videos of customized subjects is highly sought after for a wide range of emerging applications, including immersive VR/AR, virtual production, and next-generation e-commerce. However, despite rapid progress in subject-driven video generation, existing methods predominantly treat subjects as 2D entities, focusing on transferring identity through single-view visual features or textual prompts. Because real-world subjects are inherently 3D, applying these 2D-centric approaches to 3D object customization reveals a fundamental limitation: they lack the comprehensive spatial priors necessary to reconstruct the 3D geometry. Consequently, when synthesizing novel views, they must rely on generating plausible but arbitrary details for unseen regions, rather than preserving the true 3D identity. Achieving genuine 3D-aware customization remains challenging due to the scarcity of multi-view video datasets. While one might attempt to fine-tune models on limited video sequences, this often leads to temporal overfitting.

To resolve these issues, we introduce a novel framework for 3D-aware video customization, comprising \textit{3DreamBooth} and \textit{3Dapter}. \textit{3DreamBooth} decouples spatial geometry from temporal motion through a 1-frame optimization paradigm. By restricting updates to spatial representations, it effectively bakes a robust 3D prior into the model without the need for exhaustive video-based training. To enhance fine-grained textures and accelerate convergence, we incorporate \textit{3Dapter}, a visual conditioning module. Following single-view pre-training, \textit{3Dapter} undergoes multi-view joint optimization with the main generation branch via an asymmetrical conditioning strategy. This design allows the module to act as a dynamic selective router, querying view-specific geometric hints from a minimal reference set.

Our framework achieves high-fidelity, 3D-conditioned video generation while maintaining computational efficiency, as demonstrated through quantitative and qualitative evaluations.



\end{abstract}

\section{Introduction}
\label{sec:intro}
Imagine a product designer who wants to showcase a newly designed sneaker in a dynamic advertisement video, where the shoe rotates, walks on various terrains, and appears under different lighting conditions. Capturing such footage traditionally requires repeated, costly filming sessions across different environments. Ideally, one would simply capture the subject once and let a generative system handle the rest, placing it convincingly into any scene, from any viewpoint, and in any motion context. Or consider a game developer who needs to animate a custom character across dozens of diverse scenes while maintaining strict visual consistency. Such applications demand a generative system that not only understands \textit{what} a subject looks like, but also grasps its underlying 3D structure well enough to render it faithfully across unseen viewpoints and novel scenarios.

To realize such demanding applications, the generative AI community has actively explored subject-driven customization. While various structural conditioning frameworks~\cite{saharia2022palette, zhang2023adding, mou2024t2i} have been developed to control diffusion models~\cite{rombach2022high}, subject-driven customization has emerged as a particularly pivotal branch. Early optimization-based approaches pioneered this field by binding a specific subject's identity to a unique identifier~\cite{gal2022image, ruiz2023dreambooth, kumari2023multi}. However, these text-driven methods often struggle to capture high-frequency details due to the inherent information bottleneck of text embeddings. To address this, visual adapters were introduced to directly inject reference images into the diffusion process, which has become a widely adopted approach for preserving intricate structural and identity details in 2D generation~\cite{ye2023ip, li2024photomaker}.

Naturally, this customization paradigm has extended to customized Text-to-Video (T2V) generation, driven by the success of foundational video frameworks~\cite{chen2023videocrafter1, guo2023animatediff}. Recent studies have attempted to personalize video models for specific subjects or motions~\cite{wu2023tune, jiang2024videobooth, wang2024customvideo}. However, existing video customization methods predominantly rely on single-image references~\cite{chen2023videodreamer, zhao2024motiondirector} or purely textual prompts~\cite{huang2025videomage, wei2024dreamvideo}. 
Consequently, the generated subjects are inevitably bound to a rigid, 2D appearance and often fail to render consistently across drastically different and unseen viewpoints.
This limitation reveals a lack of genuine understanding regarding the subject's underlying 3D geometry. While 3D-aware spatial generation has been actively explored in the image domain~\cite{liu2023zero, shi2023mvdream}, the explicit injection of multi-view images of a 3D subject directly into video diffusion models to achieve robust 3D-consistent customization remains a largely unexplored frontier.

In this work, we present a novel framework for 3D-aware customized video generation, unifying optimization-based personalization and adapter-based conditioning. We first introduce \textit{3DreamBooth}, which fine-tunes the generative backbone via LoRA~\cite{hu2022lora} to internalize a subject’s 3D identity. Naïvely training on full video sequences often entangles spatial identity with temporal dynamics, causing the network to overfit to specific motion patterns. To avoid this, \textit{3DreamBooth} adopts a 1-frame training paradigm~\cite{wei2024dreamvideo, huang2025videomage}. By restricting inputs to single frames, temporal attention pathways are naturally bypassed, confining learning to spatial attributes while preserving the model's pre-trained motion priors.

While \textit{3DreamBooth} successfully implants the 3D identity, relying solely on a single identifier $V$ often leads to inefficient optimization and the loss of fine-grained textures. To address this, we introduce \textit{3Dapter}, a multi-view conditioning module integrated via a dual-branch architecture~\cite{zhang2025easycontrol, tan2025ominicontrol}. By utilizing LoRA, \textit{3Dapter} injects multi-view spatial features while preserving foundational weights. Following single-view pre-training for robust feature extraction~\cite{tan2025ominicontrol}, \textit{3DreamBooth} and \textit{3Dapter} are jointly fine-tuned on multi-view images. During this phase, the main branch reconstructs a target view while \textit{3Dapter} provides conditioning from a minimal set of reference views. This synergy enables highly detailed, 3D-conditioned generation while maintaining significant computational efficiency.

In summary, our main contributions are as follows:
\begin{itemize}
    \item We address the problem of 3D-aware video customization by introducing a multi-view conditioning framework that mitigates the entanglement of spatial identity and temporal dynamics in video diffusion models.
    \item We propose \textbf{\textit{3DreamBooth}}, a 1-frame optimization strategy that integrates subject-specific 3D identity into the model without requiring multi-view video datasets, effectively leveraging the inherent 3D awareness of modern video models.
    \item We introduce \textbf{\textit{3Dapter}}, a multi-view conditioning module trained through a two-stage pipeline, enabling efficient convergence and precise 3D-conditioned video generation.
    \item We introduce \textbf{\textit{3D-CustomBench}}, a curated evaluation suite for 3D-consistent video customization. Through extensive experiments, we demonstrate that our framework outperforms existing single-reference baselines and ablation variants in generating 3D-aware and identity-preserving videos.
\end{itemize}

\section{Related Works}
\label{sec:related}

\subsection{3D-Conditioned Image and Video Generation}
While single-view image conditioned diffusion models~\cite{zhang2023adding, ye2023ip, tan2025ominicontrol} have demonstrated remarkable success, conditioning diffusion models on 3D assets or multi-view images remains largely unexplored. Recently, RefAny3D~\cite{huang2026refany3d} fine-tuned FLUX~\cite{labs2025flux1kontextflowmatching} on a curated, pose-aligned object dataset to achieve 3D-asset conditioned image generation. Although promising, extending this approach to video generation remains challenging, as acquiring large-scale pose-aligned object-video pair datasets is non-trivial. Concurrently, MV-S2V~\cite{song2026mv} introduced a multi-view conditioned text-to-video framework. Unlike our proposed \textit{3DreamBooth}, MV-S2V trains the video diffusion model on large-scale synthetic video datasets with multi-view object references, resulting in significant computational costs.


\subsection{Subject-Driven Customization}
Subject-driven customization techniques aim to adapt diffusion foundation models to the user-provided objects, enabling their natural composition into diverse scenes. Early image customization methods achieved notable success through approaches such as textual binding~\cite{ruiz2023dreambooth}, textual inversion~\cite{gal2022image}, and visual adapters~\cite{ye2023ip, zhang2023adding, li2024photomaker}. With the rapid advancement of video diffusion models~\cite{videoworldsimulators2024, kong2024hunyuanvideo, wan2025wan}, these techniques have been naturally extended to the video domain, enabling a broader range of user-controllable functionalities.

Existing customization methods generally fall into two categories: (1) training-based zero-shot approaches~\cite{jiang2025vace, liu2025phantom, yuan2025identity} and (2) optimization-based approaches~\cite{wei2024dreamvideo, huang2025videomage}. The former learns to integrate visual features of a given subject, enabling rapid generation but often sacrificing fine-grained details. The latter better preserves the subject characteristics, yet their reliance on test-time optimization leads to slow inference. Our proposed framework unifies the strength of both paradigms, achieving faster optimization convergence while more effectively preserving 3D identity of the object in the synthesized video.

\begin{figure*}[t]
\centering
\includegraphics[width=\textwidth]{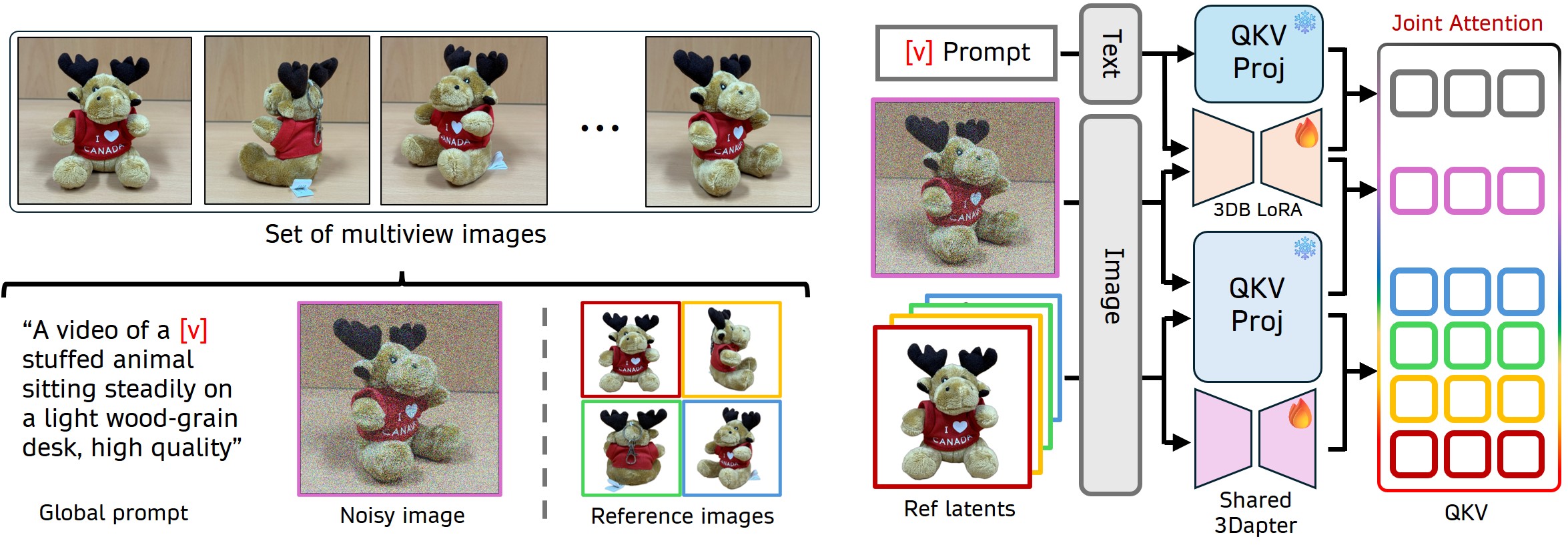}
\caption{\textbf{Overview of the 3DreamBooth training pipeline.} (Left) From multi-view images, one is selected as the target, while a sampled subset serves as reference conditions alongside a global prompt with a unique identifier $V$. (Right) The text and noisy target latents pass through the main branch (3DB LoRA), while reference latents pass through a shared \textit{3Dapter}. Their features are concatenated for Multi-view Joint Attention. This 1-frame optimization decouples spatial geometry from temporal dynamics to efficiently learn a 3D prior.}
\label{fig:training_pipeline}
\end{figure*}

\section{Method}
\label{sec:method}

\subsection{Rethinking DreamBooth for 3D Customization}
To achieve high-fidelity 3D customization of a specific subject, we build upon the foundational concept of DreamBooth~\cite{ruiz2023dreambooth}. In the image domain, DreamBooth successfully binds a unique identifier (e.g., a rare token $V$) to a specific subject by fine-tuning the model to reconstruct the subject's appearance. However, extending this concept directly to video generation models requires a deeper understanding of the interplay between spatial representation and temporal dynamics.

\subsubsection{Isolating Identity from Temporal Dynamics.} 
Typically, video diffusion models are trained on large-scale datasets to learn both appearance and motion. However, when the objective is to inject the identity of a specific subject, utilizing full video sequences of the subject for training is computationally redundant and highly prone to temporal overfitting (e.g., the model memorizing a specific motion trajectory). Based on the insight that object identity is largely a spatial attribute, we propose a 1-frame video training paradigm.

Modern video Diffusion Transformers (DiTs)~\cite{peebles2023scalable} often process inputs via joint spatio-temporal attention. While recent video customization methods typically require explicitly freezing temporal modules~\cite{huang2025videomage} or inserting separate spatial adapters to learn subject identity from static images~\cite{zhao2024motiondirector, wei2024dreamvideo}, our approach leverages an inherent architectural property: when the input is restricted to a single frame ($T=1$), the temporal attention mechanism is naturally bypassed. This effectively localizes all gradient updates exclusively to spatial representations without requiring explicit architectural modifications. Consequently, we uniquely harness this mechanism to implant the subject's comprehensive 3D visual identity into the model while implicitly preserving its pre-trained temporal priors. During inference, these untouched temporal mechanics naturally extract and drive the temporal flow of the learned identity, enabling smooth, view-consistent video generation.

\subsubsection{Eliciting the Implicit 3D Prior.} 
The core intuition behind \textit{3DreamBooth} stems from the inherent capabilities of pre-trained video diffusion models~\cite{blattmann2023stable, yang2024cogvideox, kong2024hunyuanvideo, wan2025wan}. These models already possess robust, implicit 3D priors~\cite{voleti2024sv3d, chen2024videocrafter2}. For instance, when prompted to generate a video of a ``dog'', the model naturally produces temporally coherent frames that preserve the 3D geometric consistency of the dog across different viewpoints. We hypothesize that this inherent 3D prior can be explicitly leveraged for customization.

Akin to a sculptor meticulously shaping a piece of pottery from multiple angles, we train the model using diverse static views of the target subject. Through this multi-view DreamBooth training process, the unique identifier token $V$ gradually absorbs the geometric structures and view-dependent appearances of the object. Consequently, the token $V$ evolves beyond a simple semantic identifier; it becomes a consolidated 3D prior of the specific subject. During inference, querying the model with this enriched token, combined with the pre-trained temporal dynamics, successfully yields temporally consistent videos that showcase the customized object seamlessly from arbitrary viewpoints.

\subsection{3DreamBooth}
\label{subsec:3dreambooth}
Building upon the aforementioned insights, we introduce \textit{3DreamBooth}, a novel optimization paradigm designed to inject the high-fidelity 3D identity of a specific subject into video diffusion models.

Given a set of static multi-view images of a subject, denoted as $\mathcal{S} = \{s^{(i)}\}_{i=1}^{N_\text{s}}$, where $N_\text{s}$ is the number of subject views, we treat each image as a single-frame video ($T=1$). Each view image $s^{(i)}$ is paired with an universal text prompt $p$ containing the unique identifier $V$ and a broad class noun $C$ (e.g., ``a video of a $V$ $C$''). By using a consistent prompt across all views, we force the model to internalize the multi-view spatial variations directly into the identifier token $V$, rather than relying on explicit textual view descriptions.

We optimize the pre-trained video Diffusion Transformer (DiT) $v_\theta$ using Low-Rank Adaptation (LoRA). We inject trainable weights $\phi_\text{3DB}$ into the transformer blocks (e.g., attention and MLP modules) while keeping the original model parameters, $\theta$, frozen. Since the input is restricted to $T=1$, the joint spatio-temporal attention inherently operates only across the spatial tokens. This naturally focuses the parameter updates on spatial features without disrupting the pre-trained temporal dynamics. The training objective is defined by the velocity prediction loss:
\begin{equation}
\argmin_{\phi_{\text{3DB}}} \mathbb{E}_{i, \epsilon, t} \left[ \| \bar{v} - v_{\theta, \phi_{\text{3DB}}}(z_{t}^{(i)}, t, p) \|_2^2 \right],
\end{equation}
where $i\in \{1, \dots, N_s \}$ is a sampled view index, $z_{t}^{(i)}$ is the noisy latent of a sampled view $s^{(i)}$ at diffusion timestep $t$, $\bar{v}$ represents the target velocity vector, and $p$ is the text prompt for the target subject. By minimizing this objective, the LoRA weights $\phi_{\text{3DB}}$ learn the multi-view geometric variations of the subject's appearance. Consequently, this multi-view supervision successfully bakes the comprehensive 3D identity into the token $V$ and the network weights $\phi_{\text{3DB}}$.

\begin{figure*}[t]
\centering
\includegraphics[width=\textwidth]{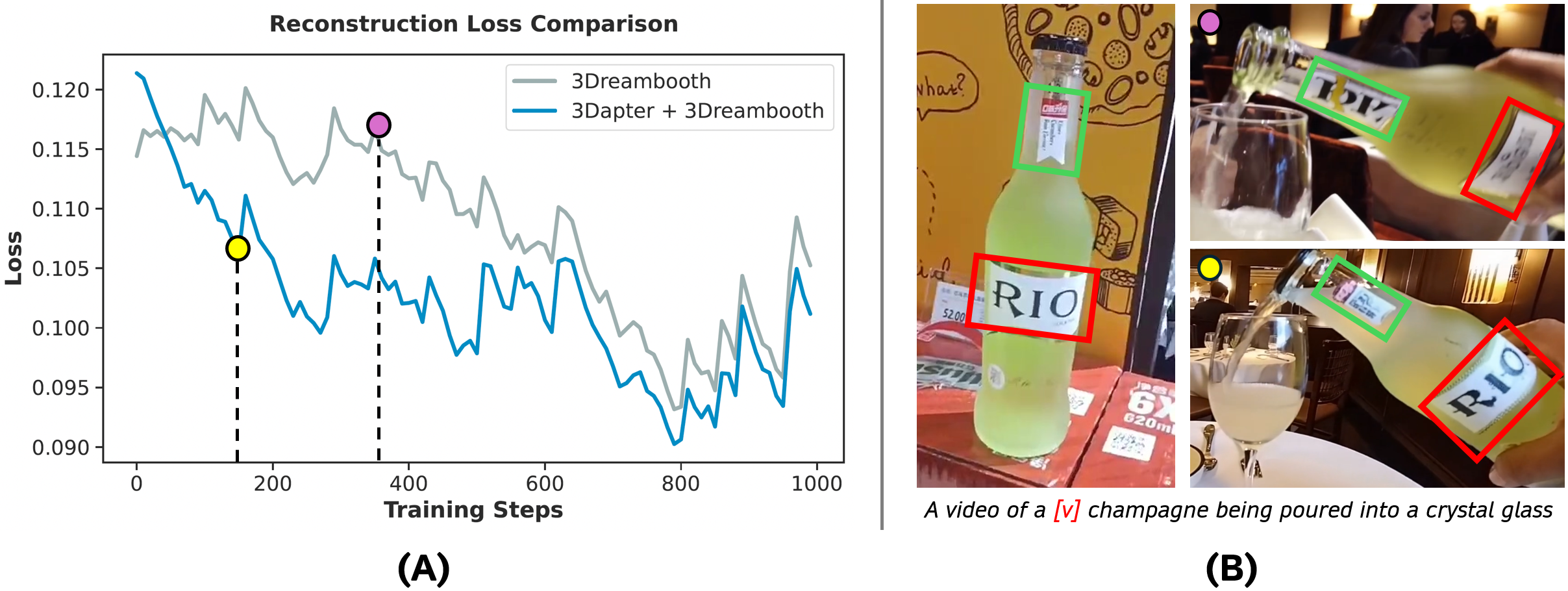}
\caption{\textbf{Convergence Analysis and Detail Preservation.} (A) Reconstruction Loss: Integrating \textit{3Dapter} (blue) drastically accelerates convergence compared to the \textit{3DreamBooth} baseline (gray). (B) Qualitative Comparison: \textit{3DreamBooth} alone (purple dot) struggles with high-frequency details due to the information bottleneck. In contrast, \textit{3Dapter}+\textit{3DreamBooth} (yellow dot) perfectly preserves intricate textures (e.g., ``RIO'' typography) much earlier, demonstrating the efficacy of explicit visual priors.}
\label{fig:loss_comparison}
\end{figure*}


\subsubsection{The Bottleneck of Text-Driven Customization.} 
While the aforementioned \textit{3DreamBooth} optimization successfully binds a 3D prior to the identifier token $V$, this text-driven approach presents two critical limitations. 
First, the optimization process is inherently slow and computationally demanding. This inefficiency arises because the model is forced to map a randomly initialized token $V$ to a complex 3D visual manifold from scratch, relying solely on text conditioning without any explicit visual hints. 
Second, and more importantly, utilizing a single token $V$ as the primary condition leads to a significant loss of fine-grained details. Although LoRA layers are optimized to memorize the subject, the text embedding is inherently designed to capture coarse semantic concepts. Consequently, the token $V$ suffers from a severe information bottleneck; it struggles to encode high-frequency details such as intricate textures, specific texts, or complex geometric nuances of the target subject.

\subsection{3Dapter}
\label{subsec:3dapter}
To overcome these limitations, recent advancements in 2D image personalization have shifted towards visual adapters~\cite{ye2023ip, mou2024t2i, li2024photomaker, wang2024instantid, guo2024pulid}, which directly inject reference images into the diffusion process to preserve both identity and intricate details.
Inspired by this paradigm, we propose \textit{3Dapter}, a multi-view conditioning module that directly injects the target subject's spatial features into the generation process.

\begin{figure*}[t]
\centering
\includegraphics[width=\textwidth]{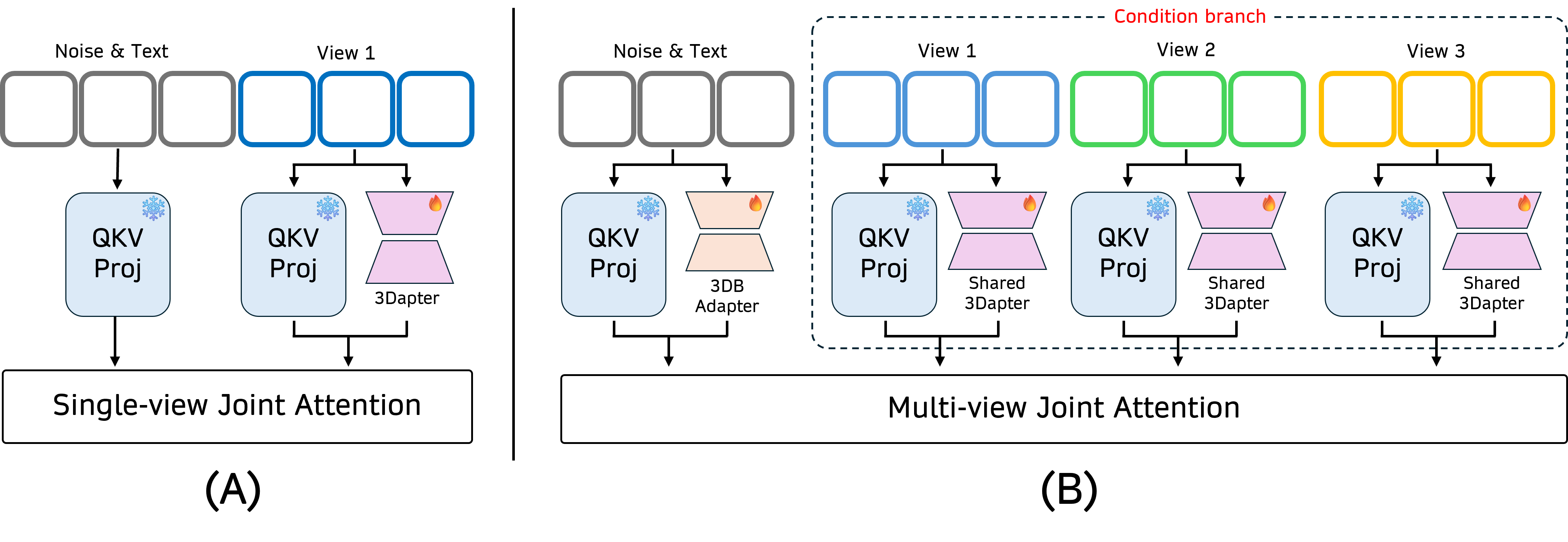}
\caption{\textbf{Detailed architecture of our two-stage conditioning mechanism.} (A) Single-view Pre-training: The visual adapter (\textit{3Dapter}) is pre-trained using single-view references and fused via Single-view Joint Attention. (B) Multi-view Joint Optimization: A trainable \textit{3DreamBooth} LoRA is added to the main branch. A minimal set of multi-view reference images is processed in parallel by the shared \textit{3Dapter}. The Multi-view Joint Attention acts as a dynamic selective router, querying relevant view-specific geometric hints to reconstruct the target view.}
\label{fig:conditioning_arch}
\end{figure*}

\subsubsection{Single-view Pre-training}
Recent advancements in controllable DiTs, such as OminiControl~\cite{tan2025ominicontrol} and EasyControl~\cite{zhang2025easycontrol}, have demonstrated the efficacy of parameter-efficient visual adapters. These frameworks typically process a condition image through a dedicated LoRA branch, concatenate the resulting condition tokens with the main generation tokens, and perform joint spatio-temporal attention.

Inspired by this paradigm, \textit{3Dapter} adopts a dual-branch forward pass for single-view conditioning, as illustrated in \cref{fig:conditioning_arch}-(A). During training, we utilize a large-scale dataset of reference-target image pairs with the corresponding text prompts, $\{(x^{(i)}, y^{(i)}, p^{(i)})\}_{i=1}^{N_D}$, where a clean reference image $x^{(i)}$ depicts a single object on a white background, whereas the target image $y^{(i)}$ shows the same object in diverse contextual scenes, accurately described by the accompanying text prompt $p^{(i)}$.
Then, we optimize the \textit{3Dapter} (LoRA) weights $\phi_\text{3Dapter}$ (while keeping the pre-trained parameters $\theta$ frozen) using the standard diffusion objective as follows:
\begin{equation}
\argmin_{\phi_{\text{3Dapter}}} = \mathbb{E}_{i, \epsilon, t}\left[ \Vert \bar{v} - v_{\theta, \phi_\text{3Dapter}}(z_t^{(i)}, t, x^{(i)}, p^{(i)}) \Vert_2^2 \right],
\label{eq:loss_3dapter}
\end{equation}
where $i\in \{1, \dots, N_D \}$ denotes a sampled dataset index, $x^{(i)}$ is a reference image, $z_t^{(i)}$ is the noisy latent of the target image $y^{(i)}$ at diffusion timestep $t$, and $p^{(i)}$ represents the corresponding text prompt.

To model the interaction between references, targets, and text prompts, we concatenate three tensors along the sequence dimension and leverage the spatio-temporal attention module in the original video model as follows:
\begin{equation}
Q = [Q_{z}, Q_{x}, Q_{p}], \quad K = [K_{y}, K_{x}, K_{p}], \quad V = [V_{y}, V_{x}, V_{p}],
\label{eq:joint_qkv}
\end{equation}
where $[\cdot, \cdot, \cdot]$ denotes concatenation, and $Q_{z}, Q_{x}, Q_{p}$ are the Query tensors from the target, reference, and the text prompt, respectively (to avoid notational clutter, we omit the layer indices of the network). $Q, K, V \in \mathbb{R}^{(2N_\text{img}+N_\text{txt}) \times d}$ are the concatenated Query, Key, and Value tensors, and $N_\text{img}$ and $N_\text{txt}$ denote the token sequence lengths of the images and the prompt text. Then, we perform the standard scaled dot-product attention using those concatenated tensors ($Q, K, V$). The reference tensors ($Q_{x}, K_{x}, V_{x}$) are produced using the respective frozen weights augmented by our trainable \textit{3Dapter} (LoRA) weights, while other tensors are generated by the frozen weights from the original video model.

\subsubsection{Multi-View Conditioning via Joint Optimization}
After single-view pre-training, we transition to the final stage, where we jointly optimize the parameters of \textit{3DreamBooth} and \textit{3Dapter} for a specific subject adaptation. Let $\mathcal{S} = \{ s^{(i)}\}_{i=1}^{N_s}$ denote the set of multi-view images of the subject. From this set, slightly abusing notation, we construct a subset of conditioning views $\mathcal{X} = \{ x^{(i)}\}_{i=1}^{N_c} $  to serve as inputs to \textit{3Dapter}, such that $\mathcal{X} \subset \mathcal{S}$ and $|\mathcal{X}|=N_c$. In addition, we preprocess the conditioning views by masking out the image backgrounds and select conditioning views that can cover the full $360^\circ$ around the subject to ensure complete spatial coverage.

Given the multi-view images of the subject $\mathcal{S}$, conditioning views $\mathcal{X}$, and an universal text prompt $p$, we simultaneously optimize the shared \textit{3Dapter} weights ($\phi_{\text{3Dapter}}$) and the \textit{3DreamBooth} weights ($\phi_{\text{3DB}}$) with the following objective,
\begin{equation}
\argmin_{\phi_{\text{3DB}}, \phi_{\text{3Dapter}}} \mathbb{E}_{i, \epsilon, t}\left[ \Vert \bar{v} - v_{ \theta, \phi_{\text{3DB}}, \phi_{\text{3Dapter}} }(z_t^{(i)}, t, \mathcal{X}, p) \Vert_2^2 \right],
\end{equation}
where $i\in \{1, \dots, N_s \}$ is a sampled view index, and $z_{t}^{(i)}$ is the noisy latent of a sampled view $s^{(i)}$ at diffusion timestep $t$.

For each joint attention module, we produce Query, Key, and Value tensors for the subject views, conditioning views, and the text prompt. For the conditioning views, we process all conditioning images $\mathcal{X}$ through a single, shared \textit{3Dapter} rather than using separate adapters for each view. This shared architecture ensures that the network extracts consistent geometric features across different viewpoints without linearly increasing the parameter count. For the subject views and the text prompt, we use the \textit{3DreamBooth} LoRA weights $\phi_{3DB}$, and all projected tensors are concatenated, as shown in \cref{fig:conditioning_arch}-(B). For example, a joint Query tensor, $Q \in \mathbb{R}^{((1+N_c) \cdot N_\text{img}+N_\text{txt}) \times d}$ can be written as follows:
\begin{equation}
Q = [Q_z, Q_{x}^{(1)}, \dots, Q_{x}^{(N_c)}, Q_p],
\end{equation}
where $[\cdot]$ denotes the concatenation operation along the sequence dimension (we also omit network layer indices for brevity). When applying 3D Rotary Positional Encoding (RoPE)~\cite{su2024roformer} to the concatenated tensors, we assign distinct, sequential temporal indices (e.g., $t=1, \dots, N_c$) to each conditioning view. This explicit temporal separation prevents the spatial features across different viewpoints from entangling, ensuring that the network processes each condition separately.

Because the \textit{3Dapter} already provides strong textural and high-frequency priors of the subject, the overall optimization burden is significantly alleviated. Rather than redundantly memorizing every intricate visual detail from scratch, the model can focus more efficiently on learning how the subject's geometry transforms across different viewpoints. This synergy results in accelerated convergence and the preservation of high-fidelity detail, effectively overcoming the inherent limitations of standard DreamBooth.

\begin{figure*}[t]
\centering
\includegraphics[width=\textwidth]{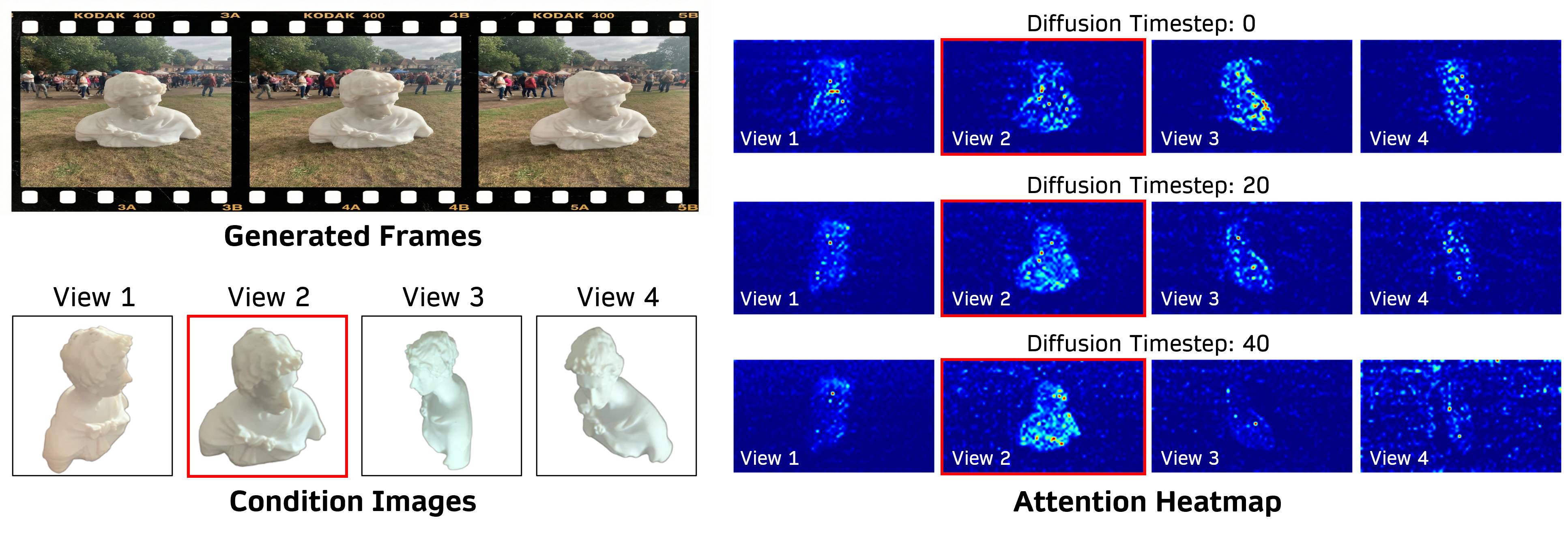}
\caption{\textbf{Visualization of the Dynamic Selective Router mechanism.} (Left) Generated frames and four multi-view conditions provided to \textit{3Dapter}. The generated poses align with View 2 (red box). (Right) Cross-attention heatmaps across diffusion timesteps ($t=0,20,40$). The network selectively assigns higher attention weights to the relevant view (View 2) to extract specific geometric features, rather than uniformly aggregating all conditions.}
\label{fig:attention_heatmap} 
\end{figure*}

We found that this design induces an interesting emergent behavior, as shown in \cref{fig:attention_heatmap}). The joint attention mechanism naturally learns to act as a dynamic, selective router. Rather than uniformly aggregating all visual features, the network explicitly learns to query and extract only the relevant, view-specific geometric hints from the concatenated multi-view reference conditions to reconstruct the current target view. By actively filtering out conflicting visual signals from irrelevant views, the network receives clean, unambiguous condition information.

\section{Experiments}
\label{sec:experiments}


\subsection{Evaluation Benchmark: 3D-CustomBench}
As 3D-aware video customization is an emerging task, there currently exists no standardized evaluation dataset. To address this, we introduce \textit{3D-CustomBench}, a novel benchmark suite designed to assess multi-view consistent video generation of customized subjects.

\noindent\textbf{Data Collection and Criteria}
The benchmark consists of 30 distinct objects selected to feature complex 3D structures, non-trivial topologies, high texture resolution, and consistent lighting. While the majority of objects are sourced from the MVImgNet~\cite{yu2023mvimgnet} dataset, we observed that its sequences often lack full $360^{\circ}$ orbital coverage. Therefore, we supplement the benchmark with our custom-captured 3D objects to cover complete $360^{\circ}$ trajectories.

\noindent\textbf{Optimization and Conditioning Strategy}
For each object, we utilize its full multi-view sequence (approximately 30 images) for \textit{3DreamBooth} optimization. We sample $N_c=4$ conditioning views for \textit{3Dapter}, selected to maximize angular coverage and minimize visual overlap.

\noindent\textbf{Validation Prompts}
We utilize GPT-4o~\cite{hurst2024gpt} to automatically generate one challenging validation prompt per object, featuring diverse backgrounds and complex dynamics (e.g., human-object interactions or environmental physics).

\subsection{Evaluation Setup and Metrics}
\label{subsec:evaluation}
To comprehensively assess our framework, we conduct extensive quantitative and qualitative comparisons against state-of-the-art subject-driven video generation baselines, VACE~\cite{jiang2025vace} and Phantom~\cite{liu2025phantom}.

\noindent\textbf{Multi-View Subject Fidelity} 
We evaluate multi-view consistency using CLIP-I, DINO-I, and an LLM-as-a-Judge~\cite{zheng2023judging} via GPT-4o~\cite{hurst2024gpt}. After generating 360-degree rotating videos and isolating the foregrounds with birefnet~\cite{zheng2024bilateral}, we compute the bi-directional maximum cosine similarity between generated frames and the four condition views using CLIP~\cite{radford2021learning} and DINOv2~\cite{oquab2023dinov2}. For human-aligned evaluation, GPT-4o assesses identity preservation across \textit{Shape}, \textit{Color}, \textit{Detail}, and \textit{Overall Identity} on a 1--5 Likert scale, averaged over five trials. Detailed protocols, mathematical definitions, and prompts are provided in the Appendix.

\noindent\textbf{3D Geometric Fidelity}
To assess 3D geometric fidelity, we employ a point cloud-based evaluation protocol. For each object in \textit{3D-CustomBench}, we reconstruct unified world-coordinate point clouds from both the ground-truth multi-view images and the generated $360^\circ$ rotation videos. Using Depth Anything 3~\cite{lin2025depth} and BiRefNet~\cite{zheng2024bilateral}, we extract per-view depth maps and foreground masks to align the generated point clouds ($\mathcal{P}_{gen}$) and the ground-truth-derived point clouds ($\mathcal{P}_{gt}$) within a shared coordinate frame. We then measure geometric consistency via Chamfer Distance~\cite{aanaes2016large}, which averages two directed nearest-neighbor distances: Accuracy ($\text{dist}(\mathcal{P}_{gen} \!\rightarrow\!\mathcal{P}_{gt})$) for shape adherence, and Completeness ($\text{dist}(\mathcal{P}_{gt} \!\rightarrow\! \mathcal{P}_{gen})$) for surface coverage. The extraction and alignment methodology, alongside the protocol diagram, are further detailed in the Appendix.

\noindent\textbf{Video Quality and Text Alignment}
Using the validation prompts from \textit{3D-CustomBench}, we evaluate the intrinsic video quality using VBench~\cite{huang2024vbench}. Specifically, we report metrics across three dimensions: \textit{Aesthetic Quality}, \textit{Imaging Quality}, and \textit{Motion Smoothness}. To measure how faithfully the generated videos reflect the text prompts, we compute a video-text alignment score using ViCLIP~\cite{wang2024internvideo2}. Additional evaluation details are provided in the Appendix.

\subsection{Implementation Details}
\label{subsec:implementation}
We build our framework upon HunyuanVideo-1.5~\cite{kong2024hunyuanvideo}, employing LoRA~\cite{hu2022lora} for both modules. \textit{3Dapter} is pre-trained for 100K iterations on Subjects200K~\cite{tan2025ominicontrol}, followed by joint optimization with \textit{3DreamBooth} for 400 iterations (learning rate $1 \times 10^{-4}$). For a fair comparison, the \textit{3DreamBooth} (only) ablation is identically optimized for 400 iterations. Full details are provided in the Appendix.

\begin{figure*}[!ht]
\centering
\includegraphics[width=\textwidth]{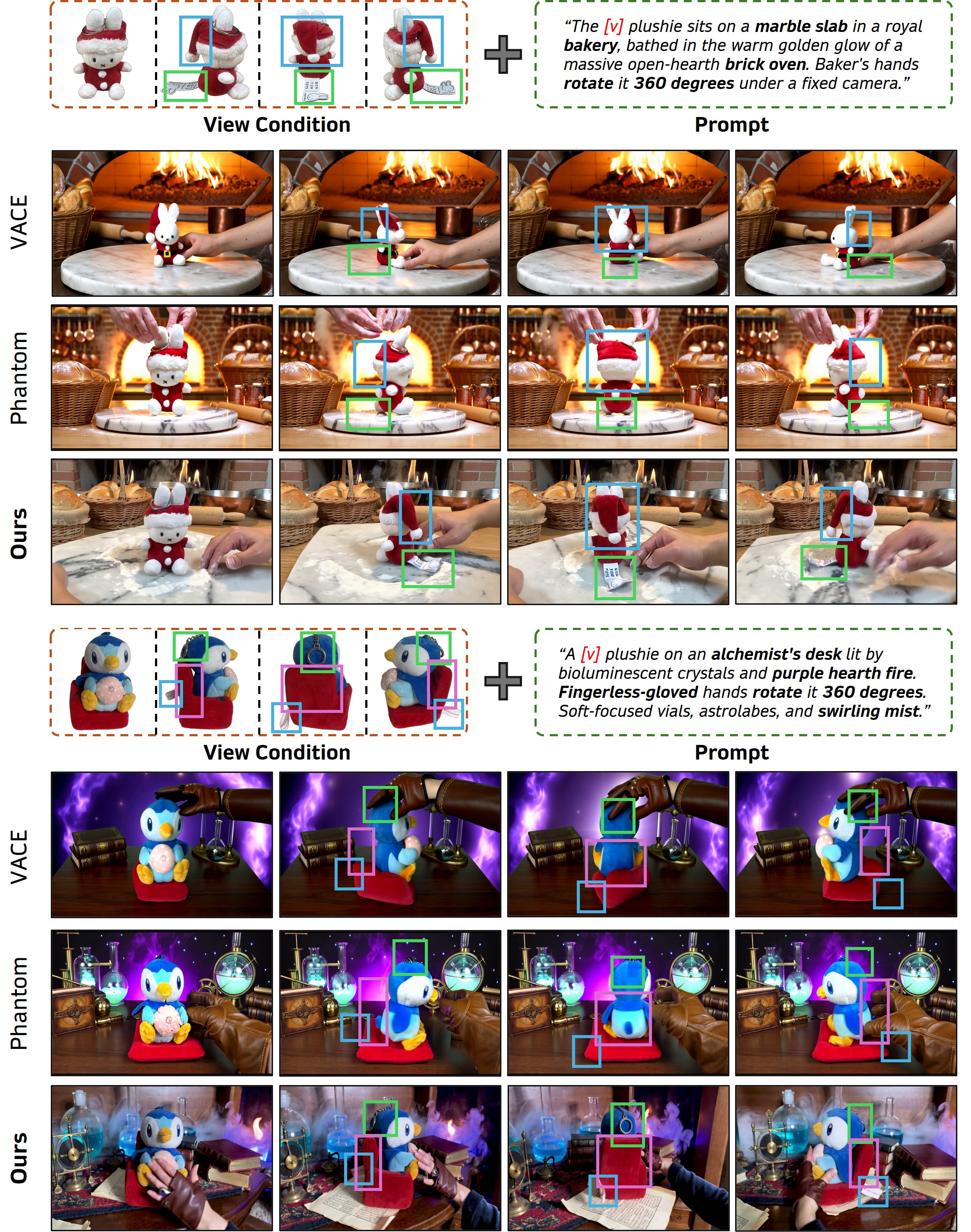}
\caption{
\textbf{Qualitative Comparisons.} We compare our proposed framework against state-of-the-art video customization baselines, VACE and Phantom. Here, ``Ours'' denotes our full synergistic model (\textit{3Dapter}+\textit{3DB}). While existing baselines struggle to reconstruct complex structures and fine-grained details (e.g., product tags and small accessories) from unseen viewpoints, our method faithfully maintains the subject's precise 3D geometry and intricate textures during dynamic $360^\circ$ rotations.
}
\label{fig:qualitative} 
\end{figure*}

\begin{table*}[t]
\centering
\caption{
\textbf{Multi-View Subject Fidelity.} We compare our framework against state-of-the-art subject-driven video generation models and our ablation variants. `S' and `M' denote Single-view and Multi-view conditioning, respectively.
}
\label{tab:multiview_consistency}
\resizebox{\columnwidth}{!}{%
\begin{tabular}{c|c|cc|cccc}
\toprule
& & \multicolumn{2}{c|}{\textbf{Feature-based}} & \multicolumn{4}{c}{\textbf{GPT-4o}} \\
\textbf{Method} & \textbf{Views} & \textbf{CLIP-I $\uparrow$} & \textbf{DINO-I $\uparrow$} & \textbf{Shape $\uparrow$} & \textbf{Color $\uparrow$} & \textbf{Detail $\uparrow$} & \textbf{Overall $\uparrow$} \\
\midrule
VACE~\cite{jiang2025vace} & S & \textbf{0.8964} & \underline{0.7395} & $\underline{4.39}_{\pm 0.05}$ & $\underline{4.09}_{\pm 0.09}$ & $\underline{3.35}_{\pm 0.15}$ & $\underline{3.95}_{\pm 0.11}$ \\
Phantom~\cite{liu2025phantom} & S & 0.8576 & 0.5861 & $3.48_{\pm 0.12}$ & $3.94_{\pm 0.13}$ & $3.03_{\pm 0.16}$ & $3.31_{\pm 0.15}$ \\
\midrule
3Dapter & S & 0.8647 & 0.5899 & $3.06_{\pm 0.03}$ & $3.09_{\pm 0.06}$ & $2.28_{\pm 0.08}$ & $2.67_{\pm 0.07}$   \\
3DreamBooth & M & 0.8382 & 0.6530 & $4.18_{\pm 0.06}$ & $3.63_{\pm 0.09}$ & $3.14_{\pm 0.11}$ & $3.53_{\pm 0.07}$ \\
\textbf{3Dapter+3DB} & M & \underline{0.8871} & \textbf{0.7420} & $\mathbf{4.80}_{\pm 0.03}$ & $\mathbf{4.53}_{\pm 0.04}$ & $\mathbf{4.04}_{\pm 0.13}$ & $\mathbf{4.57}_{\pm 0.04}$ \\
\bottomrule
\end{tabular}%
}
\end{table*}

\begin{table}[t]
\centering
\caption{
\textbf{3D Geometric Fidelity.} We compare Chamfer Distance between generated frames and ground-truth multi-view images from our \textit{3D-CustomBench}.
}
\label{tab:3d_geo}
\begin{tabular}{c|c|ccc}
\toprule
\textbf{Method} & \textbf{Views} & \textbf{Accuracy $\downarrow$} & \textbf{Completeness $\downarrow$} & \textbf{CD $\downarrow$} \\
\midrule
VACE~\cite{jiang2025vace} & S & 0.0278 & 0.0427 & 0.0353 \\
Phantom~\cite{liu2025phantom} & S & 0.0289 & 0.0388 & 0.0338 \\
\midrule
3Dapter & S & 0.0315 & 0.0659 & 0.0487 \\
3DreamBooth & M & \textbf{0.0156} & \underline{0.0322} & \underline{0.0239} \\
\textbf{3Dapter+3DB} & M & \underline{0.0182} & \textbf{0.0172} & \textbf{0.0177} \\
\bottomrule
\end{tabular}
\end{table}

\begin{table*}[t]                                              
\centering
\caption{                                                        
\textbf{Video Quality and Text Alignment.} We evaluate the intrinsic video quality using VBench~\cite{huang2024vbench} and text-video
alignment using ViCLIP Score (ViCLIP-L/14)~\cite{wang2024internvideo2}. `S' and `M' denote Single-view and Multi-view conditioning, respectively.
}
\label{tab:vbench_quality}
\resizebox{\columnwidth}{!}{%
\begin{tabular}{c|c|cccc}
\toprule
\textbf{Method} & \textbf{Type} & \textbf{Aesthetic Quality$\uparrow$} & \textbf{Imaging Quality $\uparrow$} & \textbf{Motion Smoothness $\uparrow$} &
\textbf{ViCLIP $\uparrow$} \\
\midrule
VACE~\cite{jiang2025vace} & S & 0.5915 & 70.84 & 0.9916 & \textbf{0.2663} \\
Phantom~\cite{liu2025phantom} & S & 0.5798 & 70.58 & \underline{0.9934} & \underline{0.2634} \\
\midrule
3Dapter & S & \textbf{0.6283} & 71.65 & \textbf{0.9944} & 0.2048 \\
3DreamBooth & M & 0.5245 & \underline{73.34} & 0.9928 &  0.2415 \\
\textbf{3Dapter$+$3DB} & M & \underline{0.5920} & \textbf{74.33} & 0.9918 & 0.2388 \\
\bottomrule
\end{tabular}%
}
\end{table*}

\section{Results}

\subsection{Qualitative Results}
\label{subsec:qualitative}
In Fig.~\ref{fig:qualitative}, we compare our framework against baselines, VACE~\cite{jiang2025vace} and Phantom~\cite{liu2025phantom}, using identical prompts except the subject identifier (our token $V$ versus descriptive text). Conditioned only on the first view, these methods lack spatial information for unseen regions, yielding inconsistent textures and geometries during rotation. Conversely, our framework synthesizes the full $360^{\circ}$ geometry, preserving identity across all viewpoints.

\subsection{Quantitative Results}

\noindent\textbf{Multi-View Subject Fidelity.} 
Table~\ref{tab:multiview_consistency} evaluates how well the subject's identity is preserved across views. Our full model (\textbf{3Dapter+3DB}) achieves the best performance across most metrics, particularly in GPT-4o-based human-centric evaluations (Shape, Color, Detail, Overall). Regarding the slightly higher CLIP-I score of VACE, we attribute this to CLIP's tendency to prioritize high-level semantic features over fine-grained geometric accuracy. Since VACE often generates plausible but incorrect textures for unseen views, it may achieve high semantic similarity despite failing to preserve consistent multi-view visual fidelity, which is more accurately captured by our DINO-I and GPT-4o metrics.

\noindent\textbf{3D Geometric Fidelity} 
The superiority of our framework is most evident in 3D geometric consistency (Table~\ref{tab:3d_geo}). We achieve a Chamfer Distance (CD) of 0.0177, nearly halving the error compared to the most competitive single-view method (Phantom, 0.0338). Specifically, our significant lead in Completeness (0.0172) demonstrates that our multi-view conditioning effectively recovers the full $360^{\circ}$ geometry, whereas single-view methods fail to infer the full 3D structure of unseen perspectives.

\noindent\textbf{Video Quality and Text Alignment.} As shown in Table~\ref{tab:vbench_quality}, our framework maintains high intrinsic video quality and text alignment, achieving comparable or superior performance against the baselines. Notably, we outperform existing methods in Imaging Quality while demonstrating highly competitive ViCLIP scores, ensuring that our robust 3D conditioning does not compromise the foundational generation capabilities.

\subsection{Ablation Studies}
Our ablation variants (Table~\ref{tab:multiview_consistency}, \ref{tab:3d_geo}) highlight the necessity of each component. Using \textit{3Dapter} alone (Single-view) provides strong aesthetic quality but lacks 3D consistency due to the absence of multi-view priors. Conversely, \textit{3DreamBooth} alone (Multi-view) ensures better geometry but struggles with fine-grained texture details and CLIP alignment. The joint optimization of \textit{3Dapter}+\textit{3DB} yields the best trade-off, combining the robust 3D structure of \textit{3DreamBooth} with the high-frequency feature injection of \textit{3Dapter}.


\section{Conclusion}
We introduced a highly efficient framework for 3D-aware video customization. By decoupling spatial identity from temporal motion via \textit{3DreamBooth}'s 1-frame optimization, we embed subject-specific 3D priors without temporal overfitting. Furthermore, our multi-view conditioning module, \textit{3Dapter}, acts as a dynamic selective router to explicitly extract relevant geometric features, preserving intricate textures. Extensive evaluations on our curated \textit{3D-CustomBench} demonstrate that our synergistic approach achieves state-of-the-art 3D geometric fidelity and fast convergence, paving the way for advanced applications in virtual production and advertising.



%
%
\bibliographystyle{splncs04}
\bibliography{main}

@String(ICLR  = {Int. Conf. Learn. Represent.})

@String(AAAI  = {AAAI})

@String(ICLR  = {ICLR})

@inproceedings{ruiz2023dreambooth,
  title={Dreambooth: Fine tuning text-to-image diffusion models for subject-driven generation},
  author={Ruiz, Nataniel and Li, Yuanzhen and Jampani, Varun and Pritch, Yael and Rubinstein, Michael and Aberman, Kfir},
  booktitle={Proceedings of the IEEE/CVF conference on computer vision and pattern recognition},
  pages={22500--22510},
  year={2023}
}

@inproceedings{tan2025ominicontrol,
  title={Ominicontrol: Minimal and universal control for diffusion transformer},
  author={Tan, Zhenxiong and Liu, Songhua and Yang, Xingyi and Xue, Qiaochu and Wang, Xinchao},
  booktitle={Proceedings of the IEEE/CVF International Conference on Computer Vision},
  pages={14940--14950},
  year={2025}
}

@inproceedings{zhang2025easycontrol,
  title={Easycontrol: Adding efficient and flexible control for diffusion transformer},
  author={Zhang, Yuxuan and Yuan, Yirui and Song, Yiren and Wang, Haofan and Liu, Jiaming},
  booktitle={Proceedings of the IEEE/CVF International Conference on Computer Vision},
  pages={19513--19524},
  year={2025}
}

@inproceedings{rombach2022high,
  title={High-resolution image synthesis with latent diffusion models},
  author={Rombach, Robin and Blattmann, Andreas and Lorenz, Dominik and Esser, Patrick and Ommer, Bj{\"o}rn},
  booktitle={Proceedings of the IEEE/CVF conference on computer vision and pattern recognition},
  pages={10684--10695},
  year={2022}
}

@inproceedings{mou2024t2i,
  title={T2i-adapter: Learning adapters to dig out more controllable ability for text-to-image diffusion models},
  author={Mou, Chong and Wang, Xintao and Xie, Liangbin and Wu, Yanze and Zhang, Jian and Qi, Zhongang and Shan, Ying},
  booktitle={Proceedings of the AAAI conference on artificial intelligence},
  volume={38},
  number={5},
  pages={4296--4304},
  year={2024}
}

@inproceedings{zhang2023adding,
  title={Adding conditional control to text-to-image diffusion models},
  author={Zhang, Lvmin and Rao, Anyi and Agrawala, Maneesh},
  booktitle={Proceedings of the IEEE/CVF international conference on computer vision},
  pages={3836--3847},
  year={2023}
}

@article{ye2023ip,
  title={Ip-adapter: Text compatible image prompt adapter for text-to-image diffusion models},
  author={Ye, Hu and Zhang, Jun and Liu, Sibo and Han, Xiao and Yang, Wei},
  journal={arXiv preprint arXiv:2308.06721},
  year={2023}
}

@inproceedings{jiang2024videobooth,
  title={Videobooth: Diffusion-based video generation with image prompts},
  author={Jiang, Yuming and Wu, Tianxing and Yang, Shuai and Si, Chenyang and Lin, Dahua and Qiao, Yu and Loy, Chen Change and Liu, Ziwei},
  booktitle={Proceedings of the IEEE/CVF Conference on Computer Vision and Pattern Recognition},
  pages={6689--6700},
  year={2024}
}

@inproceedings{kumari2023multi,
  title={Multi-concept customization of text-to-image diffusion},
  author={Kumari, Nupur and Zhang, Bingliang and Zhang, Richard and Shechtman, Eli and Zhu, Jun-Yan},
  booktitle={Proceedings of the IEEE/CVF conference on computer vision and pattern recognition},
  pages={1931--1941},
  year={2023}
}

@inproceedings{li2024photomaker,
  title={Photomaker: Customizing realistic human photos via stacked id embedding},
  author={Li, Zhen and Cao, Mingdeng and Wang, Xintao and Qi, Zhongang and Cheng, Ming-Ming and Shan, Ying},
  booktitle={Proceedings of the IEEE/CVF conference on computer vision and pattern recognition},
  pages={8640--8650},
  year={2024}
}

@article{guo2023animatediff,
  title={Animatediff: Animate your personalized text-to-image diffusion models without specific tuning},
  author={Guo, Yuwei and Yang, Ceyuan and Rao, Anyi and Liang, Zhengyang and Wang, Yaohui and Qiao, Yu and Agrawala, Maneesh and Lin, Dahua and Dai, Bo},
  journal={arXiv preprint arXiv:2307.04725},
  year={2023}
}

@inproceedings{wu2023tune,
  title={Tune-a-video: One-shot tuning of image diffusion models for text-to-video generation},
  author={Wu, Jay Zhangjie and Ge, Yixiao and Wang, Xintao and Lei, Stan Weixian and Gu, Yuchao and Shi, Yufei and Hsu, Wynne and Shan, Ying and Qie, Xiaohu and Shou, Mike Zheng},
  booktitle={Proceedings of the IEEE/CVF international conference on computer vision},
  pages={7623--7633},
  year={2023}
}

@inproceedings{liu2023zero,
  title={Zero-1-to-3: Zero-shot one image to 3d object},
  author={Liu, Ruoshi and Wu, Rundi and Van Hoorick, Basile and Tokmakov, Pavel and Zakharov, Sergey and Vondrick, Carl},
  booktitle={Proceedings of the IEEE/CVF international conference on computer vision},
  pages={9298--9309},
  year={2023}
}

@article{shi2023mvdream,
  title={Mvdream: Multi-view diffusion for 3d generation},
  author={Shi, Yichun and Wang, Peng and Ye, Jianglong and Long, Mai and Li, Kejie and Yang, Xiao},
  journal={arXiv preprint arXiv:2308.16512},
  year={2023}
}

@inproceedings{saharia2022palette,
  title={Palette: Image-to-image diffusion models},
  author={Saharia, Chitwan and Chan, William and Chang, Huiwen and Lee, Chris and Ho, Jonathan and Salimans, Tim and Fleet, David and Norouzi, Mohammad},
  booktitle={ACM SIGGRAPH 2022 conference proceedings},
  pages={1--10},
  year={2022}
}

@article{gal2022image,
  title={An image is worth one word: Personalizing text-to-image generation using textual inversion},
  author={Gal, Rinon and Alaluf, Yuval and Atzmon, Yuval and Patashnik, Or and Bermano, Amit H and Chechik, Gal and Cohen-Or, Daniel},
  journal={arXiv preprint arXiv:2208.01618},
  year={2022}
}

@article{chen2023videocrafter1,
  title={Videocrafter1: Open diffusion models for high-quality video generation},
  author={Chen, Haoxin and Xia, Menghan and He, Yingqing and Zhang, Yong and Cun, Xiaodong and Yang, Shaoshu and Xing, Jinbo and Liu, Yaofang and Chen, Qifeng and Wang, Xintao and others},
  journal={arXiv preprint arXiv:2310.19512},
  year={2023}
}

@inproceedings{peebles2023scalable,
  title={Scalable diffusion models with transformers},
  author={Peebles, William and Xie, Saining},
  booktitle={Proceedings of the IEEE/CVF international conference on computer vision},
  pages={4195--4205},
  year={2023}
}

@article{kong2024hunyuanvideo,
  title={Hunyuanvideo: A systematic framework for large video generative models},
  author={Kong, Weijie and Tian, Qi and Zhang, Zijian and Min, Rox and Dai, Zuozhuo and Zhou, Jin and Xiong, Jiangfeng and Li, Xin and Wu, Bo and Zhang, Jianwei and others},
  journal={arXiv preprint arXiv:2412.03603},
  year={2024}
}

@article{wan2025wan,
  title={Wan: Open and advanced large-scale video generative models},
  author={Wan, Team and Wang, Ang and Ai, Baole and Wen, Bin and Mao, Chaojie and Xie, Chen-Wei and Chen, Di and Yu, Feiwu and Zhao, Haiming and Yang, Jianxiao and others},
  journal={arXiv preprint arXiv:2503.20314},
  year={2025}
}

@article{wang2024customvideo,
  title={Customvideo: Customizing text-to-video generation with multiple subjects},
  author={Wang, Zhao and Li, Aoxue and Zhu, Lingting and Guo, Yong and Dou, Qi and Li, Zhenguo},
  journal={arXiv preprint arXiv:2401.09962},
  year={2024}
}

@inproceedings{wei2024dreamvideo,
  title={Dreamvideo: Composing your dream videos with customized subject and motion},
  author={Wei, Yujie and Zhang, Shiwei and Qing, Zhiwu and Yuan, Hangjie and Liu, Zhiheng and Liu, Yu and Zhang, Yingya and Zhou, Jingren and Shan, Hongming},
  booktitle={Proceedings of the IEEE/CVF Conference on Computer Vision and Pattern Recognition},
  pages={6537--6549},
  year={2024}
}

@inproceedings{huang2025videomage,
  title={Videomage: Multi-subject and motion customization of text-to-video diffusion models},
  author={Huang, Chi-Pin and Wu, Yen-Siang and Chung, Hung-Kai and Chang, Kai-Po and Yang, Fu-En and Wang, Yu-Chiang Frank},
  booktitle={Proceedings of the IEEE/CVF Conference on Computer Vision and Pattern Recognition},
  pages={17603--17612},
  year={2025}
}

@article{song2026mv,
  title={MV-S2V: Multi-View Subject-Consistent Video Generation},
  author={Song, Ziyang and Gong, Xinyu and Liu, Bangya and Zhao, Zelin},
  journal={arXiv preprint arXiv:2601.17756},
  year={2026}
}

@article{videoworldsimulators2024,
  title={Video generation models as world simulators},
  author={Tim Brooks and Bill Peebles and Connor Holmes and Will DePue and Yufei Guo and Li Jing and David Schnurr and Joe Taylor and Troy Luhman and Eric Luhman and Clarence Ng and Ricky Wang and Aditya Ramesh},
  year={2024},
  url={https://openai.com/research/video-generation-models-as-world-simulators},
}

@inproceedings{jiang2025vace,
  title={Vace: All-in-one video creation and editing},
  author={Jiang, Zeyinzi and Han, Zhen and Mao, Chaojie and Zhang, Jingfeng and Pan, Yulin and Liu, Yu},
  booktitle={Proceedings of the IEEE/CVF International Conference on Computer Vision},
  pages={17191--17202},
  year={2025}
}

@inproceedings{liu2025phantom,
  title={Phantom: Subject-consistent video generation via cross-modal alignment},
  author={Liu, Lijie and Ma, Tianxiang and Li, Bingchuan and Chen, Zhuowei and Liu, Jiawei and Li, Gen and Zhou, Siyu and He, Qian and Wu, Xinglong},
  booktitle={Proceedings of the IEEE/CVF International Conference on Computer Vision},
  pages={14951--14961},
  year={2025}
}

@inproceedings{yuan2025identity,
  title={Identity-preserving text-to-video generation by frequency decomposition},
  author={Yuan, Shenghai and Huang, Jinfa and He, Xianyi and Ge, Yunyang and Shi, Yujun and Chen, Liuhan and Luo, Jiebo and Yuan, Li},
  booktitle={Proceedings of the Computer Vision and Pattern Recognition Conference},
  pages={12978--12988},
  year={2025}
}

@inproceedings{chen2024videocrafter2,
  title={Videocrafter2: Overcoming data limitations for high-quality video diffusion models},
  author={Chen, Haoxin and Zhang, Yong and Cun, Xiaodong and Xia, Menghan and Wang, Xintao and Weng, Chao and Shan, Ying},
  booktitle={Proceedings of the IEEE/CVF conference on computer vision and pattern recognition},
  pages={7310--7320},
  year={2024}
}

@inproceedings{voleti2024sv3d,
  title={Sv3d: Novel multi-view synthesis and 3d generation from a single image using latent video diffusion},
  author={Voleti, Vikram and Yao, Chun-Han and Boss, Mark and Letts, Adam and Pankratz, David and Tochilkin, Dmitry and Laforte, Christian and Rombach, Robin and Jampani, Varun},
  booktitle={European Conference on Computer Vision},
  pages={439--457},
  year={2024},
  organization={Springer}
}

@article{blattmann2023stable,
  title={Stable video diffusion: Scaling latent video diffusion models to large datasets},
  author={Blattmann, Andreas and Dockhorn, Tim and Kulal, Sumith and Mendelevitch, Daniel and Kilian, Maciej and Lorenz, Dominik and Levi, Yam and English, Zion and Voleti, Vikram and Letts, Adam and others},
  journal={arXiv preprint arXiv:2311.15127},
  year={2023}
}

@article{yang2024cogvideox,
  title={Cogvideox: Text-to-video diffusion models with an expert transformer},
  author={Yang, Zhuoyi and Teng, Jiayan and Zheng, Wendi and Ding, Ming and Huang, Shiyu and Xu, Jiazheng and Yang, Yuanming and Hong, Wenyi and Zhang, Xiaohan and Feng, Guanyu and others},
  journal={arXiv preprint arXiv:2408.06072},
  year={2024}
}

@inproceedings{zhao2024motiondirector,
  title={Motiondirector: Motion customization of text-to-video diffusion models},
  author={Zhao, Rui and Gu, Yuchao and Wu, Jay Zhangjie and Zhang, David Junhao and Liu, Jia-Wei and Wu, Weijia and Keppo, Jussi and Shou, Mike Zheng},
  booktitle={European Conference on Computer Vision},
  pages={273--290},
  year={2024},
  organization={Springer}
}

@article{hu2022lora,
  title={Lora: Low-rank adaptation of large language models.},
  author={Hu, Edward J and Shen, Yelong and Wallis, Phillip and Allen-Zhu, Zeyuan and Li, Yuanzhi and Wang, Shean and Wang, Liang and Chen, Weizhu and others},
  journal={Iclr},
  volume={1},
  number={2},
  pages={3},
  year={2022}
}

@article{chen2023videodreamer,
  title={Videodreamer: Customized multi-subject text-to-video generation with disen-mix finetuning},
  author={Chen, Hong and Wang, Xin and Zeng, Guanning and Zhang, Yipeng and Zhou, Yuwei and Han, Feilin and Zhu, Wenwu},
  journal={arXiv preprint arXiv:2311.00990},
  year={2023}
}

@misc{labs2025flux1kontextflowmatching,
      title={FLUX.1 Kontext: Flow Matching for In-Context Image Generation and Editing in Latent Space},
      author={Black Forest Labs and Stephen Batifol and Andreas Blattmann and Frederic Boesel and Saksham Consul and Cyril Diagne and Tim Dockhorn and Jack English and Zion English and Patrick Esser and Sumith Kulal and Kyle Lacey and Yam Levi and Cheng Li and Dominik Lorenz and Jonas Müller and Dustin Podell and Robin Rombach and Harry Saini and Axel Sauer and Luke Smith},
      year={2025},
      eprint={2506.15742},
      archivePrefix={arXiv},
      primaryClass={cs.GR},
      url={https://arxiv.org/abs/2506.15742},
}

@inproceedings{yu2023mvimgnet,
  title={Mvimgnet: A large-scale dataset of multi-view images},
  author={Yu, Xianggang and Xu, Mutian and Zhang, Yidan and Liu, Haolin and Ye, Chongjie and Wu, Yushuang and Yan, Zizheng and Zhu, Chenming and Xiong, Zhangyang and Liang, Tianyou and others},
  booktitle={Proceedings of the IEEE/CVF conference on computer vision and pattern recognition},
  pages={9150--9161},
  year={2023}
}

@article{zheng2024bilateral,
  title={Bilateral reference for high-resolution dichotomous image segmentation},
  author={Zheng, Peng and Gao, Dehong and Fan, Deng-Ping and Liu, Li and Laaksonen, Jorma and Ouyang, Wanli and Sebe, Nicu},
  journal={arXiv preprint arXiv:2401.03407},
  year={2024}
}

@article{lin2025depth,
  title={Depth anything 3: Recovering the visual space from any views},
  author={Lin, Haotong and Chen, Sili and Liew, Junhao and Chen, Donny Y and Li, Zhenyu and Shi, Guang and Feng, Jiashi and Kang, Bingyi},
  journal={arXiv preprint arXiv:2511.10647},
  year={2025}
}

@inproceedings{rusu2009fast,
  title={Fast point feature histograms (FPFH) for 3D registration},
  author={Rusu, Radu Bogdan and Blodow, Nico and Beetz, Michael},
  booktitle={2009 IEEE international conference on robotics and automation},
  pages={3212--3217},
  year={2009},
  organization={IEEE}
}

@article{fischler1981random,
  title={Random sample consensus: a paradigm for model fitting with applications to image analysis and automated cartography},
  author={Fischler, Martin A and Bolles, Robert C},
  journal={Communications of the ACM},
  volume={24},
  number={6},
  pages={381--395},
  year={1981},
  publisher={ACM New York, NY, USA}
}

@inproceedings{besl1992method,
  title={Method for registration of 3-D shapes},
  author={Besl, Paul J and McKay, Neil D},
  booktitle={Sensor fusion IV: control paradigms and data structures},
  volume={1611},
  pages={586--606},
  year={1992},
  organization={Spie}
}

@article{umeyama2002least,
  title={Least-squares estimation of transformation parameters between two point patterns},
  author={Umeyama, Shinji},
  journal={IEEE Transactions on pattern analysis and machine intelligence},
  volume={13},
  number={4},
  pages={376--380},
  year={2002},
  publisher={IEEE}
}

@article{aanaes2016large,
  title={Large-scale data for multiple-view stereopsis},
  author={Aan{\ae}s, Henrik and Jensen, Rasmus Ramsb{\o}l and Vogiatzis, George and Tola, Engin and Dahl, Anders Bjorholm},
  journal={International Journal of Computer Vision},
  volume={120},
  number={2},
  pages={153--168},
  year={2016},
  publisher={Springer}
}

@inproceedings{huang2024vbench,
  title={Vbench: Comprehensive benchmark suite for video generative models},
  author={Huang, Ziqi and He, Yinan and Yu, Jiashuo and Zhang, Fan and Si, Chenyang and Jiang, Yuming and Zhang, Yuanhan and Wu, Tianxing and Jin, Qingyang and Chanpaisit, Nattapol and others},
  booktitle={Proceedings of the IEEE/CVF Conference on Computer Vision and Pattern Recognition},
  pages={21807--21818},
  year={2024}
}

@inproceedings{radford2021learning,
  title={Learning transferable visual models from natural language supervision},
  author={Radford, Alec and Kim, Jong Wook and Hallacy, Chris and Ramesh, Aditya and Goh, Gabriel and Agarwal, Sandhini and Sastry, Girish and Askell, Amanda and Mishkin, Pamela and Clark, Jack and others},
  booktitle={International conference on machine learning},
  pages={8748--8763},
  year={2021},
  organization={PmLR}
}

@article{oquab2023dinov2,
  title={Dinov2: Learning robust visual features without supervision},
  author={Oquab, Maxime and Darcet, Timoth{\'e}e and Moutakanni, Th{\'e}o and Vo, Huy and Szafraniec, Marc and Khalidov, Vasil and Fernandez, Pierre and Haziza, Daniel and Massa, Francisco and El-Nouby, Alaaeldin and others},
  journal={arXiv preprint arXiv:2304.07193},
  year={2023}
}

@article{hurst2024gpt,
  title={Gpt-4o system card},
  author={Hurst, Aaron and Lerer, Adam and Goucher, Adam P and Perelman, Adam and Ramesh, Aditya and Clark, Aidan and Ostrow, AJ and Welihinda, Akila and Hayes, Alan and Radford, Alec and others},
  journal={arXiv preprint arXiv:2410.21276},
  year={2024}
}

@article{zheng2023judging,
  title={Judging llm-as-a-judge with mt-bench and chatbot arena},
  author={Zheng, Lianmin and Chiang, Wei-Lin and Sheng, Ying and Zhuang, Siyuan and Wu, Zhanghao and Zhuang, Yonghao and Lin, Zi and Li, Zhuohan and Li, Dacheng and Xing, Eric and others},
  journal={Advances in neural information processing systems},
  volume={36},
  pages={46595--46623},
  year={2023}
}

@article{wang2024instantid,
  title={Instantid: Zero-shot identity-preserving generation in seconds},
  author={Wang, Qixun and Bai, Xu and Wang, Haofan and Qin, Zekui and Chen, Anthony and Li, Huaxia and Tang, Xu and Hu, Yao},
  journal={arXiv preprint arXiv:2401.07519},
  year={2024}
}

@article{guo2024pulid,
  title={Pulid: Pure and lightning id customization via contrastive alignment},
  author={Guo, Zinan and Wu, Yanze and Zhuowei, Chen and Zhang, Peng and He, Qian and others},
  journal={Advances in neural information processing systems},
  volume={37},
  pages={36777--36804},
  year={2024}
}

@article{su2024roformer,
  title={Roformer: Enhanced transformer with rotary position embedding},
  author={Su, Jianlin and Ahmed, Murtadha and Lu, Yu and Pan, Shengfeng and Bo, Wen and Liu, Yunfeng},
  journal={Neurocomputing},
  volume={568},
  pages={127063},
  year={2024},
  publisher={Elsevier}
}

@inproceedings{wang2024internvideo2,
  title={Internvideo2: Scaling foundation models for multimodal video understanding},
  author={Wang, Yi and Li, Kunchang and Li, Xinhao and Yu, Jiashuo and He, Yinan and Chen, Guo and Pei, Baoqi and Zheng, Rongkun and Wang, Zun and Shi, Yansong and others},
  booktitle={European conference on computer vision},
  pages={396--416},
  year={2024},
  organization={Springer}
}

@inproceedings{huang2026refany3d,
  title={RefAny3D: 3D Asset-Referenced Diffusion Models for Image Generation},
  author={Huang, Hanzhuo and Bao, Qingyang and Gu, Zekai and Du, Zhongshuo and Lin, Cheng and Liu, Yuan and Yang, Sibei},
  booktitle={The Fourteenth International Conference on Learning Representations},
  year={2026}
}

\clearpage
\appendix

\renewcommand{\thefigure}{A\arabic{figure}}
\setcounter{figure}{0}


\begin{center}
  \Large \textbf{Supplementary Material: \\ 3DreamBooth: High-Fidelity 3D Subject-Driven Video Generation Model}
\end{center}
\vspace{2em}

\section{Additional Qualitative Results}
\label{sec:supp_qualitative}

To further demonstrate the robust 3D-aware generation capabilities of our proposed framework, we provide additional qualitative results extending beyond the examples shown in the main paper.

\begin{figure*}[!ht]
\centering
\includegraphics[width=\textwidth]{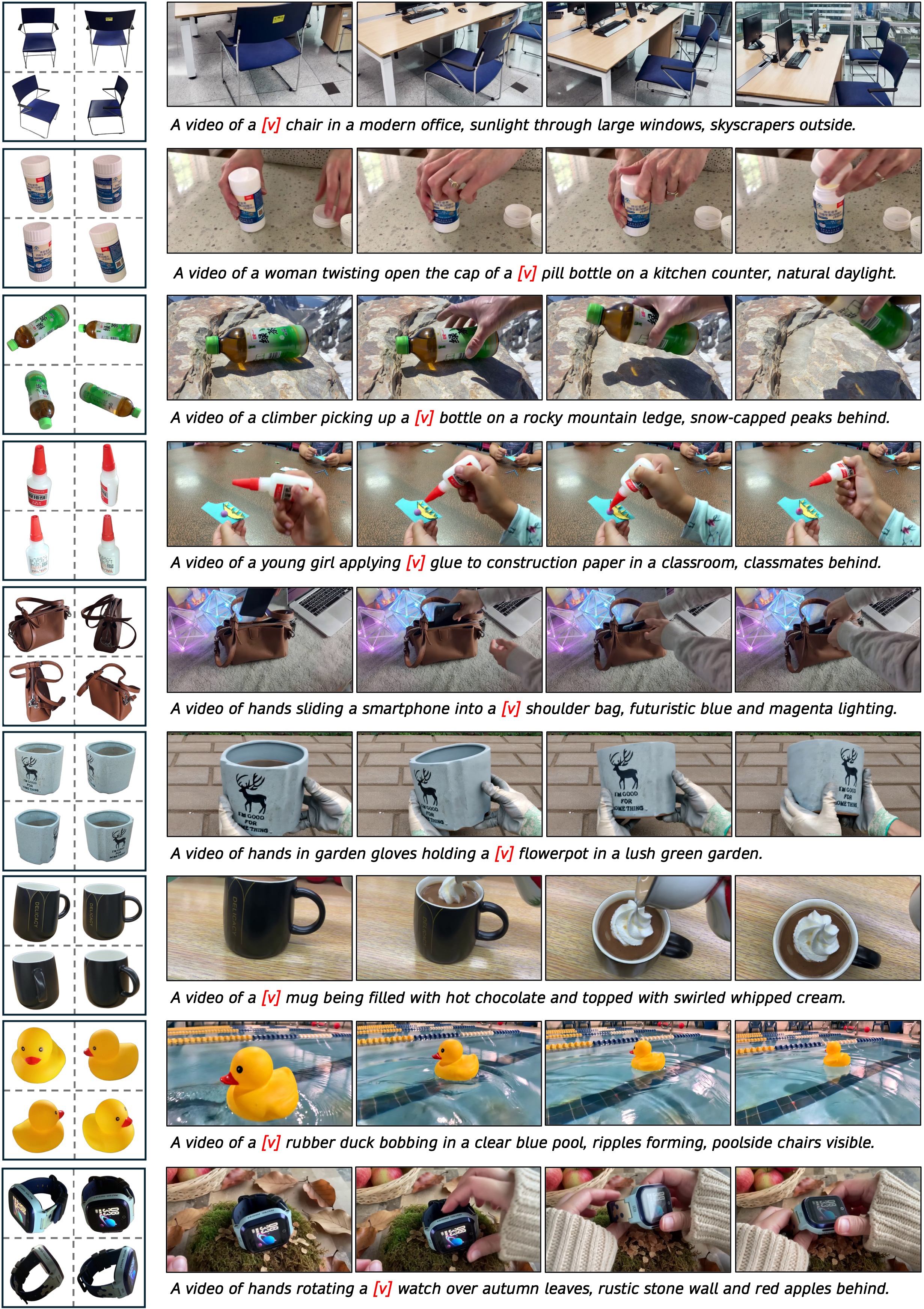}
\caption{\textbf{Qualitative results across diverse subjects.} Each row presents a distinct object, demonstrating accurate 3D identity preservation in dynamic contexts.}
\label{fig:supple_quali_one} 
\end{figure*}

\begin{figure*}[!ht]
\centering
\includegraphics[width=\textwidth]{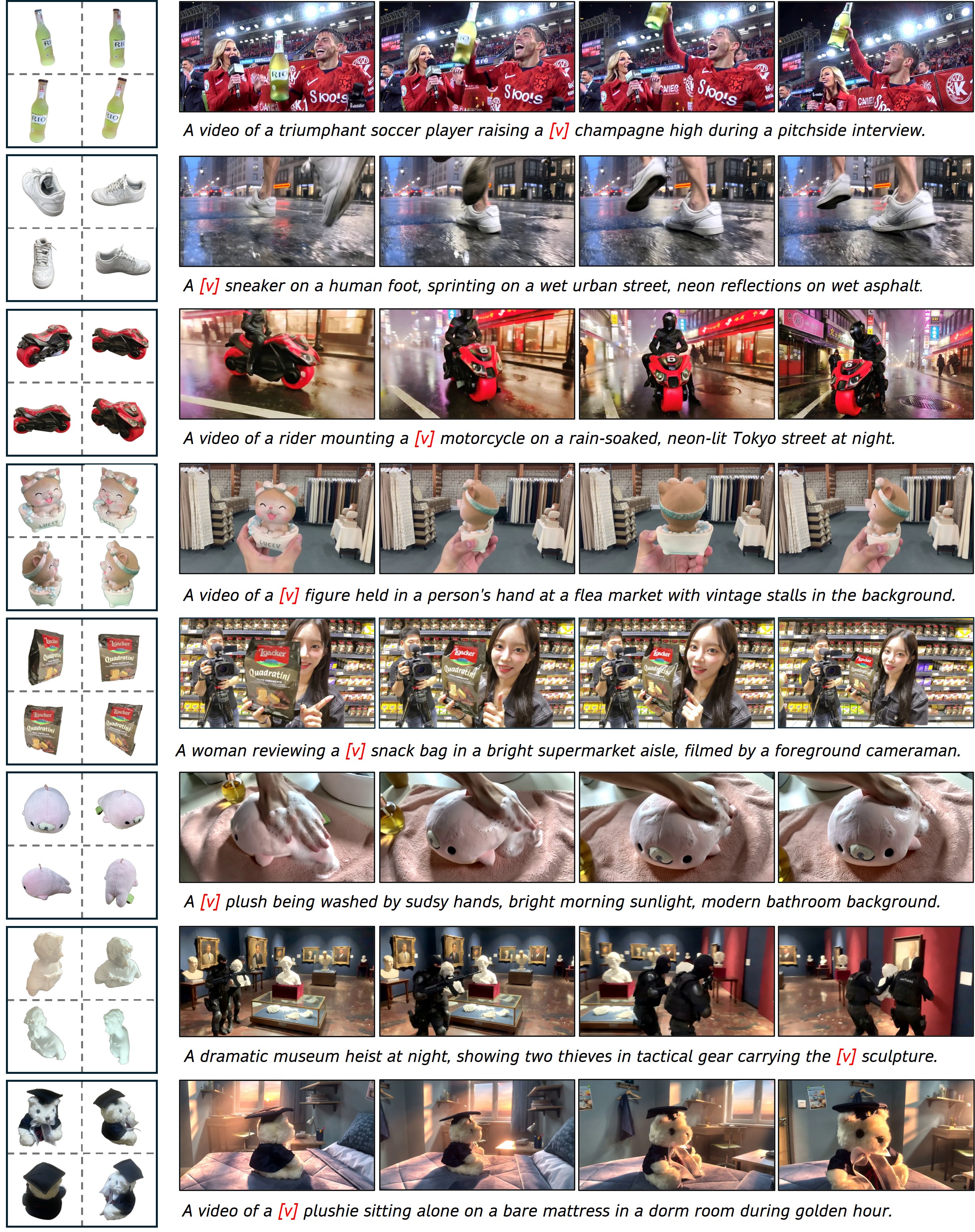}
\caption{\textbf{Additional qualitative results across diverse subjects.} 
Each row presents a distinct object, further demonstrating the 
generalizability of our framework in preserving 3D identity across 
varied object categories and dynamic contexts.}
\label{fig:supple_quali_two} 
\end{figure*}

\begin{figure*}[!ht]
\centering
\includegraphics[width=\textwidth]{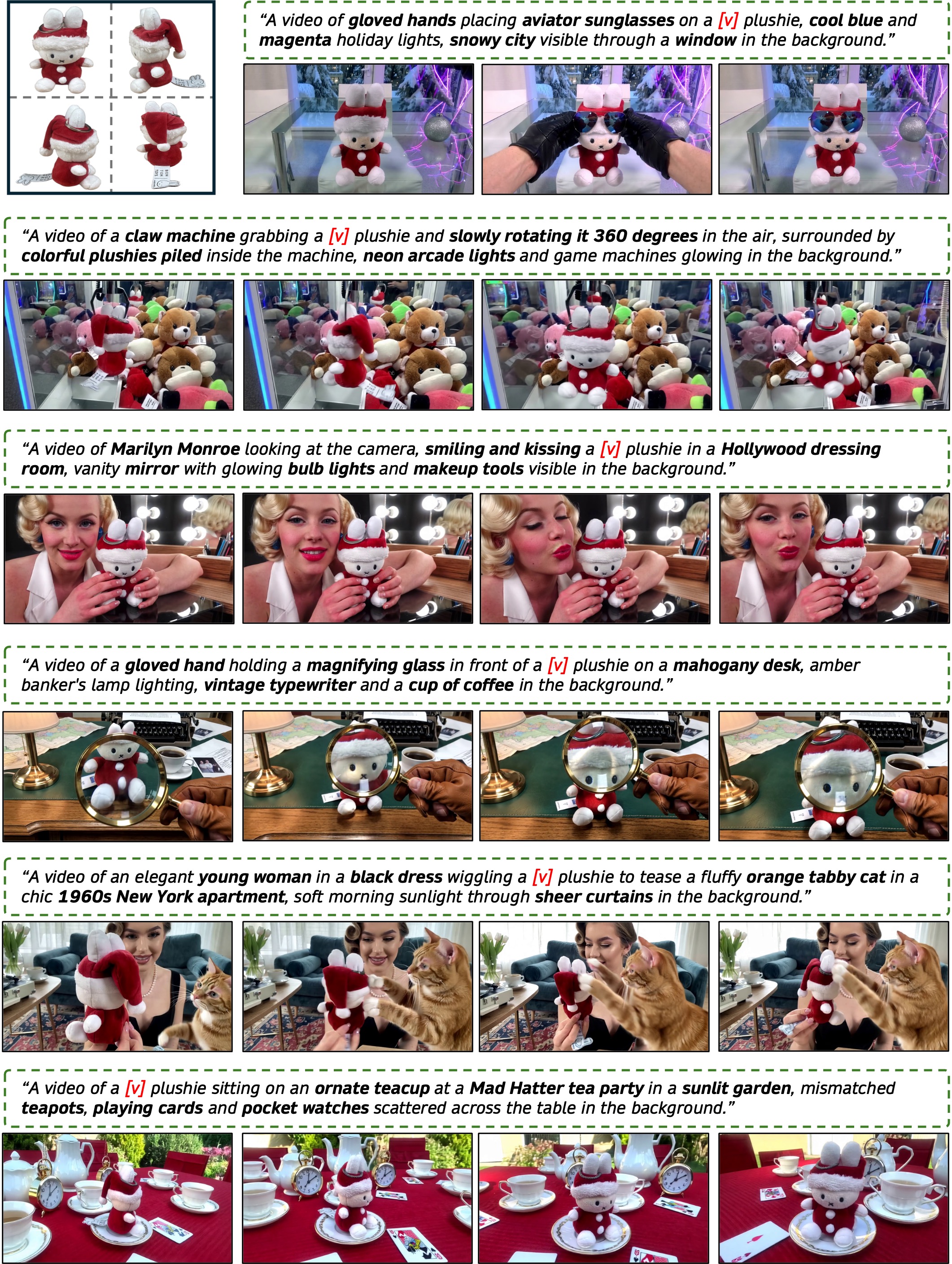}
\caption{\textbf{Single-subject generation across diverse prompts.} Given a single subject (top-left), we generate videos using multiple prompts with varying scenarios. Our framework consistently preserves the subject's identity across complex scenes, including human-object interactions and richly detailed environments.}

\label{fig:supple_quali_second} 
\end{figure*}

\noindent\textbf{Diverse Subject Generation.} In Figs.~\ref{fig:supple_quali_one} and~\ref{fig:supple_quali_two}, we present generated videos across a wide variety of object categories. For each subject, our \textit{3Dapter}\allowbreak+\allowbreak\textit{3DreamBooth} joint optimization framework effectively bakes the robust 3D spatial prior into the generation process. As a result, the generated videos faithfully preserve the intricate 3D identity, high-frequency textures, and complex geometric structures of the subjects, even when seamlessly integrated into highly dynamic environmental contexts.

\noindent\textbf{Robustness Across Diverse Scenarios.} Furthermore, Fig.~\ref{fig:supple_quali_second} illustrates the versatility of our framework when a single customized subject is driven by multiple, distinct text prompts. Despite the drastic variations in the background scenes and the presence of challenging human-object interactions, our model consistently maintains subject identity without any structural degradation. This confirms that our framework preserves the generative capabilities of the pre-trained video model while achieving consistent subject customization.

\section{Ablation Study on 3Dapter Single-View Pre-training}
\label{sec:supp_ablation_pretrain}

To validate the critical necessity of the single-view pre-training stage for the \textit{3Dapter} module, we conduct a visual ablation study. A fundamental question in our framework is whether the initial single image-to-video (I2V) pre-training is strictly required, or if the network can learn visual conditioning entirely from scratch during the joint optimization phase.

\begin{figure}[!t]
    \centering
    \includegraphics[width=\textwidth]{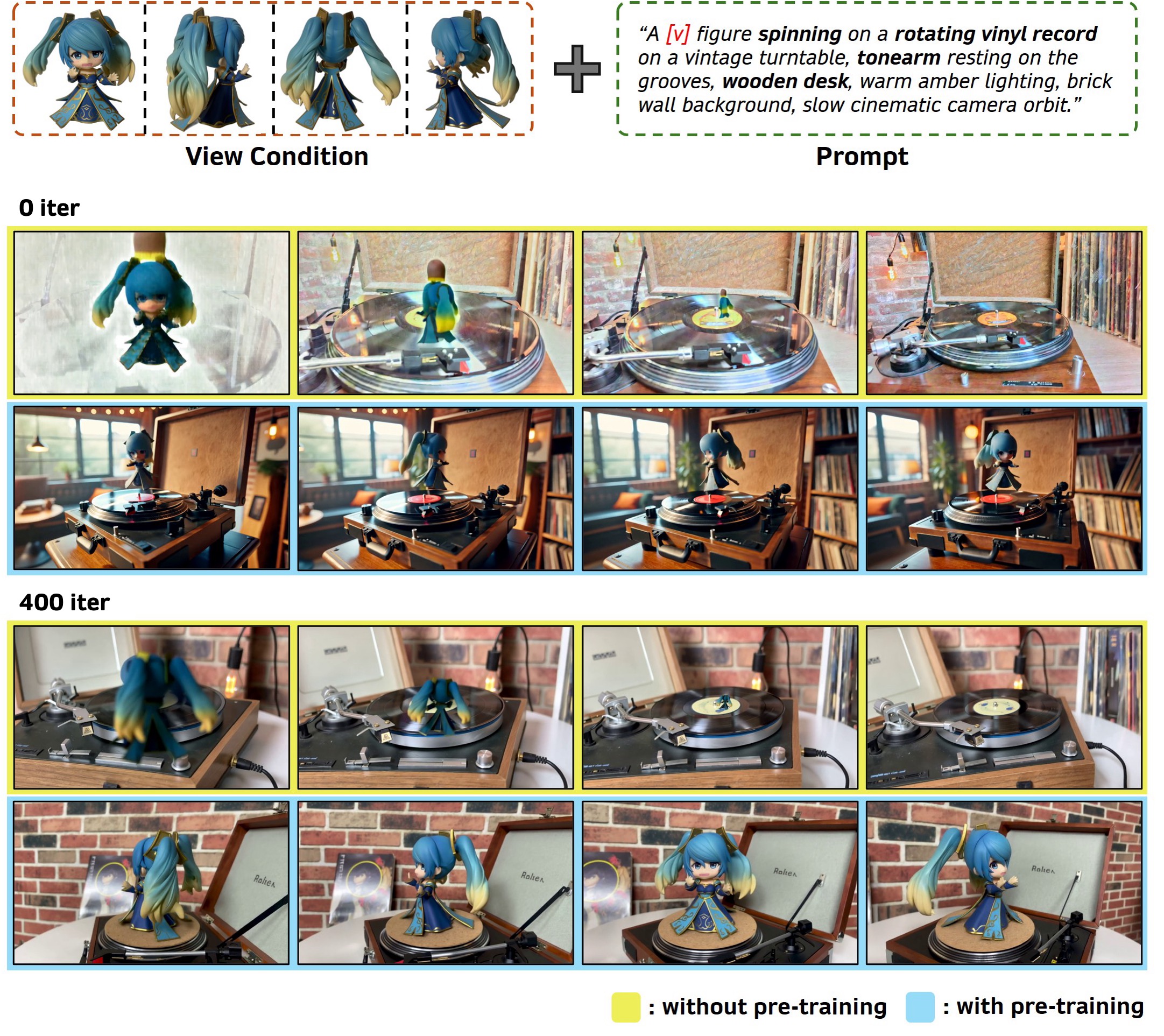}
    \caption{\textbf{Effect of \textit{3Dapter} pre-training on joint optimization.} Randomly initialized \textit{3Dapter} (yellow borders) versus pre-trained \textit{3Dapter} (blue borders) across different \textit{3DreamBooth} joint training iterations. Without pre-training, the model suffers optimization collapse even at 400 iterations. In contrast, the pre-trained \textit{3Dapter} enables accurate multi-view alignment and high-fidelity 3D customization within a short optimization period.}
    \label{fig:supp_ablation_pretrain}
\end{figure}

We compare the generated videos using a randomly initialized \textit{3Dapter} (without pre-training) against our fully pre-trained \textit{3Dapter} (with pre-training) at 0 and 400 iterations of \textit{3Dapter}+\textit{3DreamBooth} joint optimization. As demonstrated in Fig.~\ref{fig:supp_ablation_pretrain}, without pre-training (yellow borders), the randomly initialized module lacks any fundamental visual conditioning capability. At 0 iterations, it produces meaningless noise. More importantly, even after 400 iterations of joint optimization, the model completely fails to converge. Forcing the network to simultaneously learn basic image-to-video mapping and complex multi-view spatial alignment from scratch causes severe optimization collapse, resulting in structural artifacts and a loss of identity.

In contrast, the fully pre-trained \textit{3Dapter} (blue borders) provides a robust structural foundation. At 0 iterations, the multi-view reference frames are processed as a sequential signal under 3D RoPE (temporal+spatial) positional encoding, causing rotation artifacts. However, through joint optimization, the model learns to selectively attend to the most relevant multi-view cues, effectively acting as a spatial router that extracts precise 3D structural information from the reference views. This enables accurate and high-fidelity 3D-aware video customization within just 400 iterations.

\section{Training Dynamics and Qualitative Ablation}
\label{sec:supp_ablation_dynamics}

To comprehensively validate the necessity and synergistic effect of our proposed modules, we conduct an extended analysis focusing on both the training dynamics and a comparative qualitative ablation.

\begin{figure*}[!t]
\centering
\includegraphics[width=\textwidth]{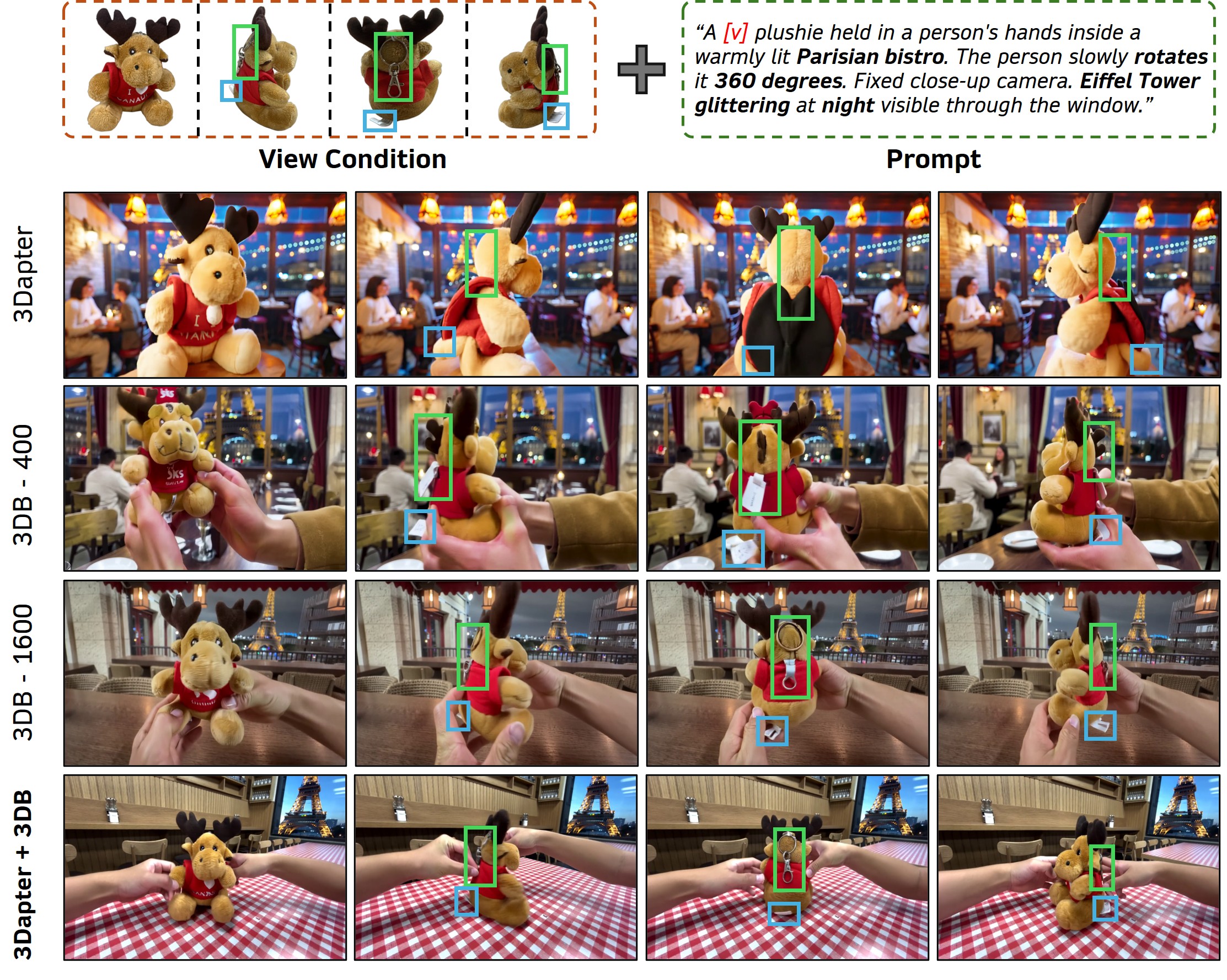}
\caption{\textbf{Comprehensive qualitative ablation across three configurations.} 
Comparing \textit{3Dapter} only, \textit{3DreamBooth} only (at 400 and 1600 iterations), and our full framework at 400 iterations (\textit{3Dapter}+\textit{3DreamBooth}). Compared to our full model (\textit{3Dapter}+\textit{3DreamBooth}), \textit{3Dapter} alone fails to maintain 3D geometric volume and view consistency, while \textit{3DreamBooth} alone lacks structural precision and texture fidelity even with $4\times$ more iterations.}
\label{fig:supp_ablation_moose} 
\end{figure*}

\begin{figure*}[!t]
\centering
\includegraphics[width=\textwidth]{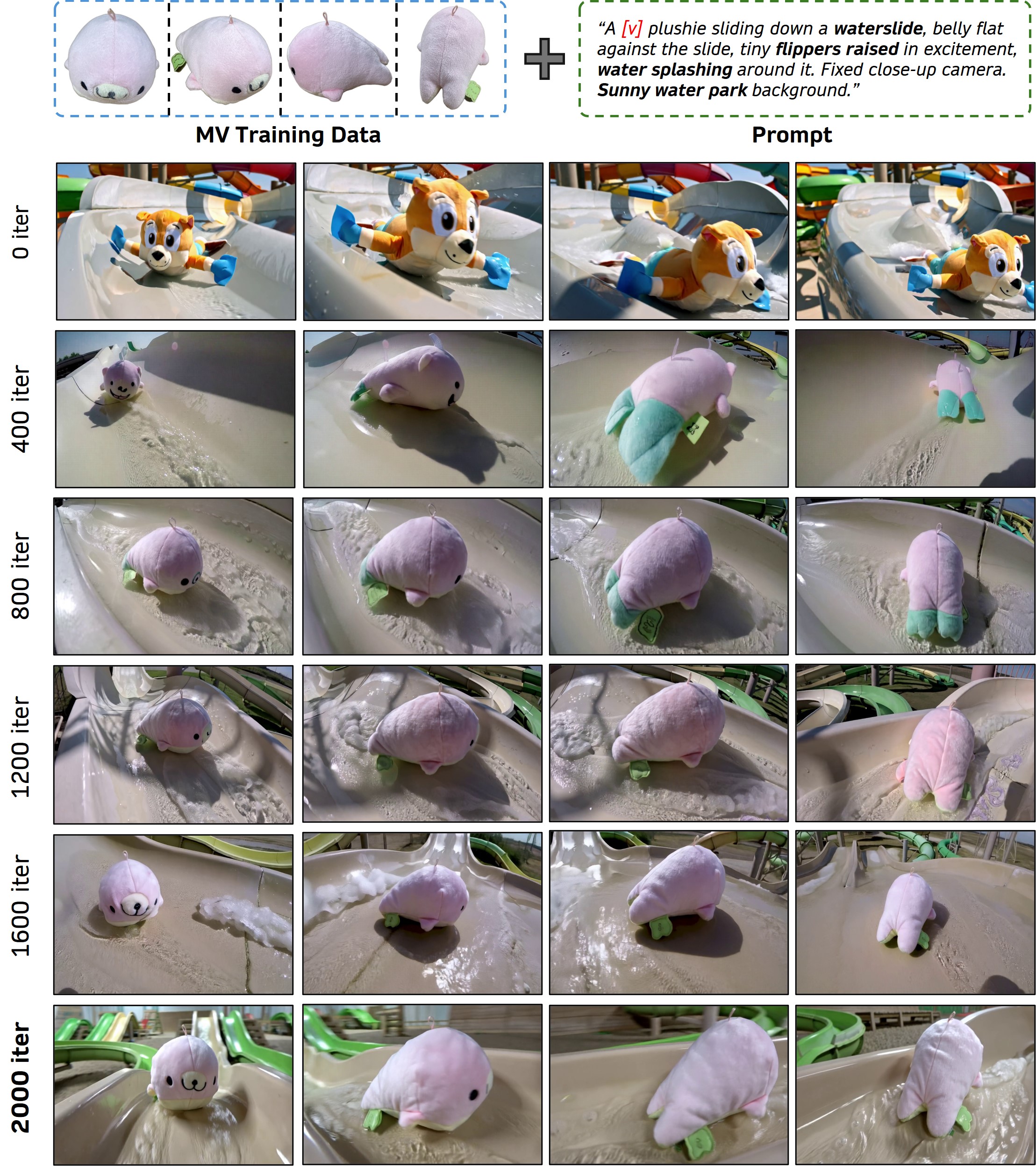}
\caption{\textbf{Step-by-step training dynamics of the \textit{3DreamBooth}-only baseline.} 
Generated frames at 0, 400, 800, 1200, 1600, and 2000 training iterations illustrate the convergence dynamics of the \textit{3DreamBooth}-only baseline. While the coarse shape and color of the reference plushie are visible from early iterations, accurate subject identity and view-consistent geometry emerge only with prolonged optimization.}
\label{fig:supp_step_convergence} 
\end{figure*}

\subsection{Comprehensive Qualitative Ablation}
Fig.~\ref{fig:supp_ablation_moose} presents a comprehensive visual ablation study that complements the convergence analysis. We compare three configurations: \textit{3Dapter} only, \textit{3DreamBooth} only (at 400 and 1600 iterations), and our full framework \textit{3Dapter}\allowbreak+\allowbreak\textit{3DreamBooth}.

When deploying \textit{3Dapter} alone, the model fails to guarantee accurate 3D spatial volume and view consistency, as zero-shot visual conditioning struggles to build a robust 3D prior. Conversely, while the fully converged \textit{3DreamBooth} (1600 iterations) maintains multi-view consistency, it still falls short of capturing the exact structural precision and crisp textures. In contrast, our full framework (\textit{3Dapter}+\textit{3DreamBooth}), benefiting from the explicit multi-view visual hints provided by \textit{3Dapter}, seamlessly integrates precise spatial conditions with a robust 3D volume. It achieves exceptional identity and detail preservation within just 400 iterations. This comprehensive comparison supports that our dynamic selective routing mechanism not only drastically accelerates the convergence of 3D customization but is also indispensable for high-fidelity detail preservation.

\subsection{3DreamBooth Training Dynamics}
As discussed in the main text, relying solely on text-driven optimization (\textit{3DreamBooth}) introduces an information bottleneck, leading to a notably slow convergence rate. To explicitly demonstrate this, we analyze the step-by-step training dynamics of the \textit{3DreamBooth} (only) baseline. As shown in Fig.~\ref{fig:supp_step_convergence}, we extract generated frames at 400, 800, 1200, 1600, and 2000 training iterations. At 400 iterations, while the baseline captures the coarse spatial volume of the subject, it struggles to reconstruct high-frequency details (e.g., intricate textures, logos, or typography). Although prolonged training up to 2000 iterations gradually refines the subject's appearance, fine-grained details such as textures and typography remain poorly reconstructed, indicating a fundamental limitation of text-driven optimization alone.

\section{Extensibility of 3DreamBooth to Other DiT Architectures}
\label{sec:qualitative}

\begin{figure*}[!t]
\centering
\includegraphics[width=\textwidth]{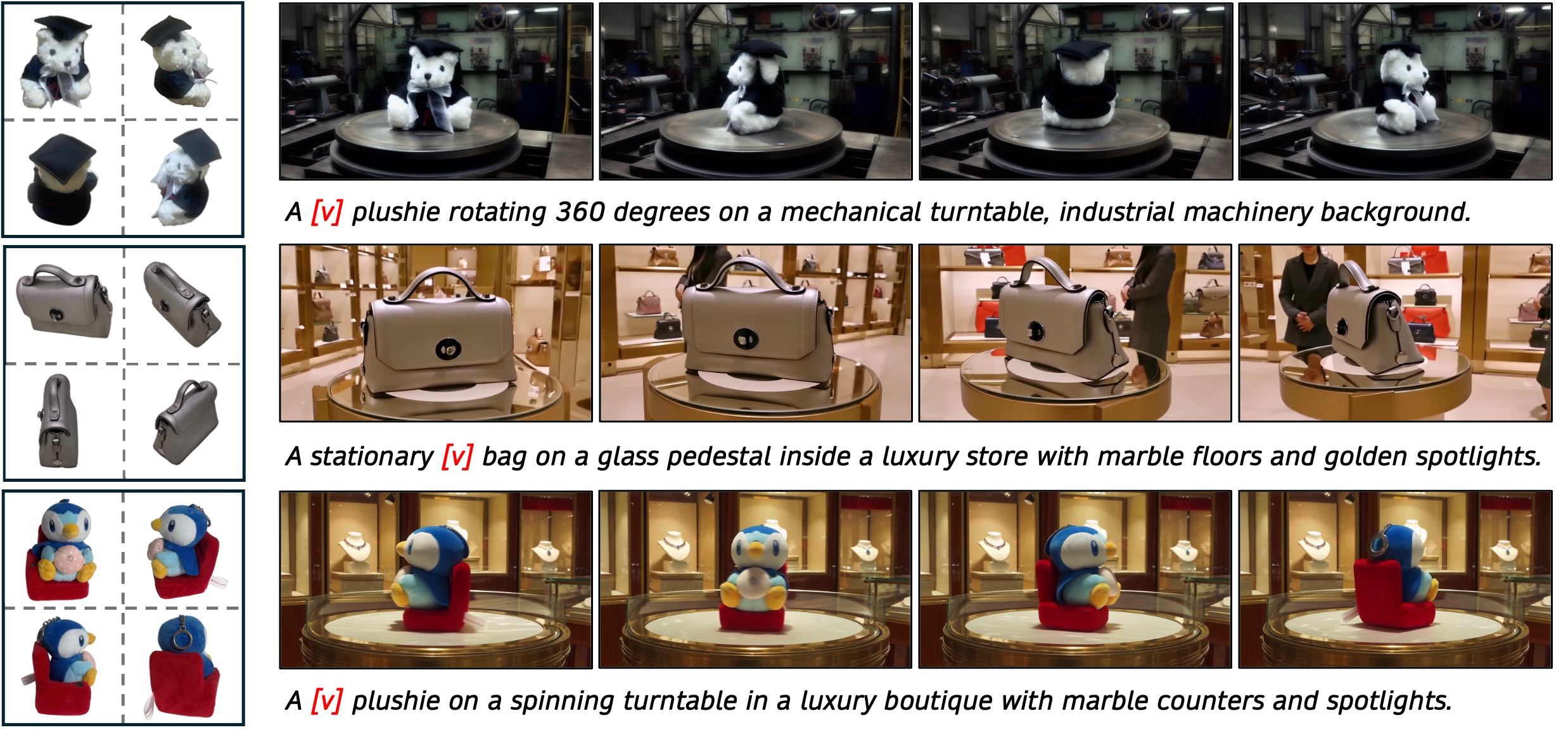}
\caption{\textbf{Extension to DiT-based Video Models.} We demonstrate the extensibility of \textit{3DreamBooth} to state-of-the-art Diffusion Transformer (DiT) architectures by applying our fine-tuning paradigm to the Wan 2.1 text-to-video model. Even without explicit spatial conditioning, the generated sequences faithfully preserve the intricate 3D identity and complex geometric structures of the subjects across diverse contexts.}
\label{fig:qualitative.jpg} 
\end{figure*}

To demonstrate the generalizability and robustness of our core optimization strategy, we investigate whether the one-frame training paradigm of \textit{3DreamBooth} can be effectively applied to other state-of-the-art Diffusion Transformer (DiT)-based video models. To isolate and explicitly evaluate the extensibility of this fine-tuning paradigm, we apply \textit{3DreamBooth} independently, without the explicit spatial conditioning of the \textit{3Dapter} module. We implement this on the widely adopted Wan 2.1 text-to-video (T2V) model~\cite{wan2025wan}.

As illustrated in Fig.~\ref{fig:qualitative.jpg}, even when relying solely on the \textit{3DreamBooth} optimization paradigm without external spatial priors, the framework successfully binds the complex multi-view identity to the unique identifier $V$. The generated results demonstrate that the multi-view spatial priors are successfully injected into the identifier and accurately reflected in the synthesized videos. This confirms that our fundamental one-frame optimization approach is highly extensible and can adapt to diverse, state-of-the-art backbone architectures.

\section{Discussion and Limitations}
\label{sec:supp_discussion_limitations}

\noindent\textbf{Inefficiency of Repurposing Task-Specific Baselines.} 
An intuitive alternative to our framework might be directly applying the \textit{3DreamBooth} multi-view fine-tuning to existing zero-shot DiT models (e.g., VACE or Phantom) without a decoupled conditioning branch. However, these task-specific models entangle video generation and visual conditioning within a single unified pathway. Fine-tuning the entire heavily parameterized model using a multi-view reconstruction objective is not only profoundly inefficient but also unstable, as it risks corrupting the core video generation capabilities. In practice, attempting this brute-force joint training easily leads to Out-of-Memory (OOM) failures. To contextualize the computational burden, training baselines like VACE required 128 A100 GPUs, and Phantom reported utilizing over 30,000 A100 GPU hours. In contrast, our decoupled \textit{3Dapter} module was efficiently pre-trained in just 4 days using only 4 RTX Pro 6000 GPUs, proving that our modular approach is fundamentally more sustainable and efficient.

\noindent\textbf{Test-Time Optimization vs. Zero-Shot Generation.} 
One might point to the requirement of test-time optimization as a limitation compared to pure zero-shot methods. However, achieving true zero-shot 3D-aware video customization would require an unimaginably massive dataset of strictly aligned multi-view and video pairs. Currently, such datasets do not exist, and curating them from real-world data is prohibitively difficult. Furthermore, even if such data were available, the required training scale would be astronomical. We elegantly circumvent this bottleneck via our efficient one-frame optimization paradigm. More importantly, in practical real-world applications (e.g., the advertising industry or commercial content creation), the absolute preservation of fine-grained subject details is paramount. For such use cases, the minor trade-off of a short test-time optimization is vastly outweighed by the uncompromising, high-fidelity 3D structural accuracy that our method guarantees over zero-shot approximations.

\noindent\textbf{Semantic Interactions and Future Work.}
Our qualitative results reveal a compelling emergent property: through \textit{3DreamBooth} training, the unique identifier $V$ successfully inherits the full spectrum of semantic interactions natively associated with the prior class noun $C$. If the class noun can perform an action, the customized subject $V$ $C$ can replicate it flawlessly within the generated video.
However, our current experiments primarily focus on rigid or static objects. It remains an open question how this paradigm adapts to highly dynamic subjects with complex articulations (e.g., human bodies) or objects undergoing drastic state changes over time. Exploring these temporal state variations represents an exciting direction for future research. Additionally, extending this optimization framework to video editing models (leveraging reference videos) could unlock highly robust, 3D-aware subject insertion in real-world footage, further broadening the impact of our approach.

\section{Ablation on Visual Conditioning and Convergence}
\label{sec:supp_ablation_convergence}

\begin{figure*}[t]
\centering
\includegraphics[width=\textwidth]{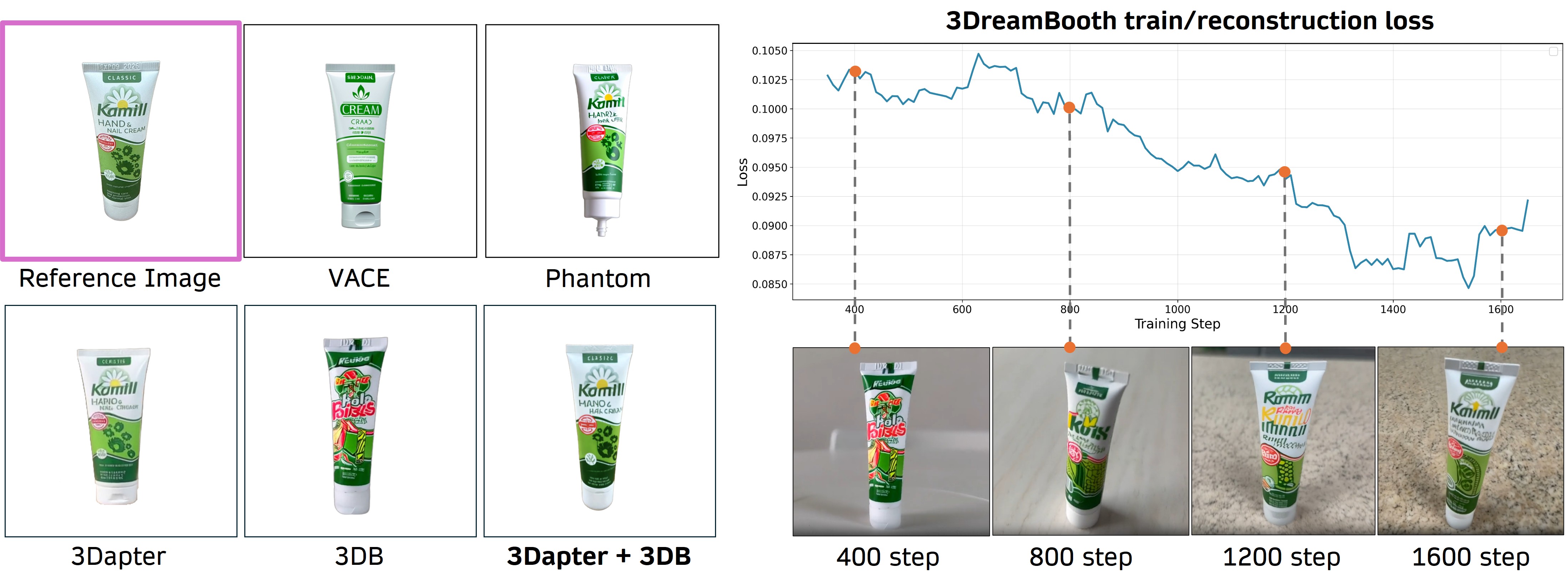}
\caption{\textbf{Visual comparison and convergence analysis.} 
(Left) Ablation study on the components of our framework. 
Single-view baselines (VACE, Phantom) and individual modules (3Dapter, 3DreamBooth) struggle to maintain fine-grained identity details, whereas the synergistic combination of \textit{3Dapter}+\textit{3DreamBooth} successfully achieves accurate identity preservation. (Right) Training loss curve and qualitative progression of \textit{3DreamBooth} across optimization steps, demonstrating that the model requires a prolonged optimization process to fully converge and refine subject details.}
\label{fig:ablation_handcream} 
\end{figure*}

To comprehensively validate the necessity of our dual-branch design, we perform a visual ablation study comparing our framework against existing baselines and isolated modules. As illustrated in the left panel of Fig.~\ref{fig:ablation_handcream}, zero-shot video customization baselines (VACE and Phantom) struggle to preserve fine-grained high-frequency details, such as the precise typography and logo on the target subject. Furthermore, as discussed in the main paper and elaborated in the subsequent qualitative sections, these baselines—along with deploying \textit{3Dapter} alone—fail to maintain accurate 3D spatial volume and view consistency. Focusing back on the single-view detail preservation in Fig.~\ref{fig:ablation_handcream}, relying solely on \textit{3DreamBooth} for 400 optimization steps captures the coarse geometry but produces garbled text due to the inherent information bottleneck of text embeddings. In contrast, our synergistic framework (\textit{3Dapter}+\textit{3DreamBooth}) successfully achieves highly accurate identity and detail preservation.

Furthermore, the right panel of Fig.~\ref{fig:ablation_handcream} provides a detailed convergence analysis of the \textit{3DreamBooth} (only) baseline. The reconstruction loss curve and the corresponding generated frames across extended optimization steps (from 400 to 1600) reveal that relying purely on text-driven optimization requires a significantly prolonged training period to refine intricate textures. Even at 1600 steps, the generated text remains less crisp compared to our full framework's result achieved at just 400 steps. Additional qualitative video comparisons are provided in Figs.~\ref{fig:supp_ablation_moose} and~\ref{fig:supp_step_convergence}.

\section{Implementation Details}
\label{sec:supp_implementation}

\subsection{Base Model and LoRA Configurations}
We build our framework upon the state-of-the-art open-source video diffusion model, HunyuanVideo-1.5~\cite{kong2024hunyuanvideo}. To efficiently inject the 3D multi-view prior without disrupting the pre-trained spatio-temporal dynamics, we employ Low-Rank Adaptation (LoRA)~\cite{hu2022lora} for both the \textit{3Dapter} and the \textit{3DreamBooth} modules. Specifically, we set the LoRA rank to 16 and the alpha scaling factor to 32 across all transformer blocks (e.g., attention and MLP modules). For inference, we generate 720p videos of 81 frames with 50 denoising steps. Notably, baseline methods such as VACE~\cite{jiang2025vace} and Phantom~\cite{liu2025phantom} are built upon WAN~\cite{wan2025wan} (14B parameters) and require large-scale task-specific retraining of the entire backbone. In contrast, our framework builds upon HunyuanVideo-1.5~\cite{kong2024hunyuanvideo} (8.3B parameters) and trains only ${\sim}$95.62M parameters (1.15\% of the backbone) via LoRA, without any modification to the pre-trained weights. This highlights the parameter efficiency of our adaptation strategy.

\subsection{3Dapter Pre-training and Dataset Curation}
For the single-view pre-training of \textit{3Dapter}, we utilize the Subjects200K dataset \cite{tan2025ominicontrol}. Curated specifically for subject-driven generation tasks, this dataset effectively addresses the overfitting risks associated with using identical image pairs by providing natural variations in pose, lighting, and contextual backgrounds. The automated construction pipeline begins with GPT-4o~\cite{hurst2024gpt} generating over 30,000 distinct subject descriptions across multiple scenarios. These descriptions are systematically formulated into structured prompts for the FLUX.1 model~\cite{labs2025flux1kontextflowmatching}, which synthesizes paired images that share the exact subject identity but differ in their contextual layout. To guarantee high fidelity, GPT-4o acts as an automated evaluator to discard any misaligned samples. During the single-view pre-training phase, we leverage this identity-consistent dataset to train \textit{3Dapter} to robustly extract and inject subject-specific spatial features into the generation process using only a single reference image. To achieve parameter-efficient learning while effectively integrating these visual conditions, we apply Low-Rank Adaptation (LoRA) \cite{hu2022lora} to the key image-processing modules of the architecture. Specifically, the trainable LoRA weights, configured with a rank of $r=16$ and a scaling parameter of $\alpha=32$, are injected into the initial image projection layer, all query, key, value, and output projection matrices within the attention blocks, as well as the fully-connected layers of the multi-layer perceptrons (MLPs). The module is optimized for $100,000$ iterations with a global batch size of 4 (1 per GPU) and a learning rate of $1 \times 10^{-4}$ using the AdamW optimizer. This pre-training phase was conducted on 4 NVIDIA RTX PRO 6000 GPUs and took approximately 4 days.

\subsection{Test-Time Joint Optimization and Hardware Setup}
During the test-time joint optimization phase for each subject in our \textit{3D-Custom\-Bench}, we explicitly preprocess all reference multi-view images to remove backgrounds using the \texttt{birefnet-massive} model~\cite{zheng2024bilateral}. This preprocessing step ensures that the visual distribution of the test-time references is consistent with the background-free condition images from the Subjects200K dataset used during the \textit{3Dapter} pre-training phase, reducing the domain gap and allowing the network to focus on the subject's 3D geometry without background interference. For the multi-view joint optimization of both the \textit{3Dapter} and \textit{3DreamBooth} modules, we maintain the same LoRA hyperparameters ($r=16$, $\alpha=32$) utilized during pre-training. While \textit{3Dapter} retains its previously defined target modules, the LoRA weights for \textit{3DreamBooth} are systematically injected into both image and text processing pathways. Specifically, these target modules encompass the input projection layers, the query, key, value, and output projection matrices across both text and image attention blocks, and their corresponding MLPs. Empirically, we found that applying LoRA to the text processing branches is particularly crucial for successful personalization; we hypothesize that this enables the model to effectively bind the unique identifier $V$ to the subject's highly specific 3D geometric features within the attention space. The joint optimization process is conducted for $400$ iterations with a batch size of 1 and a learning rate of $1 \times 10^{-4}$. All subject-specific test-time optimizations were performed on a single NVIDIA RTX PRO 6000 GPU, requiring approximately 13 minutes per subject.

\section{Detailed Evaluation Protocols}
\label{sec:supp_evaluation}

\subsection{Multi-View Subject Fidelity and LLM-as-a-Judge}
\label{subsec:multiview_fidelity}
For the multi-view subject fidelity evaluation, we first generate a 
standardized $360$-degree rotation video for each object in our 
\textit{3D-CustomBench} using the following prompt template:

\medskip
\noindent\textit{``A video of a $V$ $C$ centered against a pure, solid 
white background. The camera smoothly orbits 360 degrees, keeping the 
entire object fully in frame, high quality''}
\medskip

\noindent where $V$ denotes the learned unique identifier token and $C$ 
is replaced with the corresponding class noun for each subject 
(e.g., ``plushie'', ``mug'').
 The plain white 
background is deliberately chosen to isolate the subject from 
environmental distractors, ensuring that all downstream metrics capture 
subject fidelity rather than scene-level artifacts. All videos are 
generated at 81 frames with 50 denoising steps. For baseline methods that do not employ a learned identifier token 
(e.g., VACE~\cite{jiang2025vace}, Phantom~\cite{liu2025phantom}), 
we replace $V$ with a concise visual description 
of the subject generated by GPT-4o~\cite{hurst2024gpt} 
(e.g., ``red ceramic mug with a deer illustration'').

For feature-based multi-view consistency evaluation, we utilize the \texttt{[CLS]} tokens from pre-trained CLIP~\cite{radford2021learning} and DINOv2~\cite{oquab2023dinov2} models. Because the camera orbits the subject in our generated videos, a valid generated frame should strongly align with at least one specific condition view. Thus, we compute the cosine similarity between each generated frame and the four condition views, select the maximum similarity score for that frame, and average these maximums across the entire video.

\begin{figure*}[t]
\centering
\includegraphics[width=\textwidth]{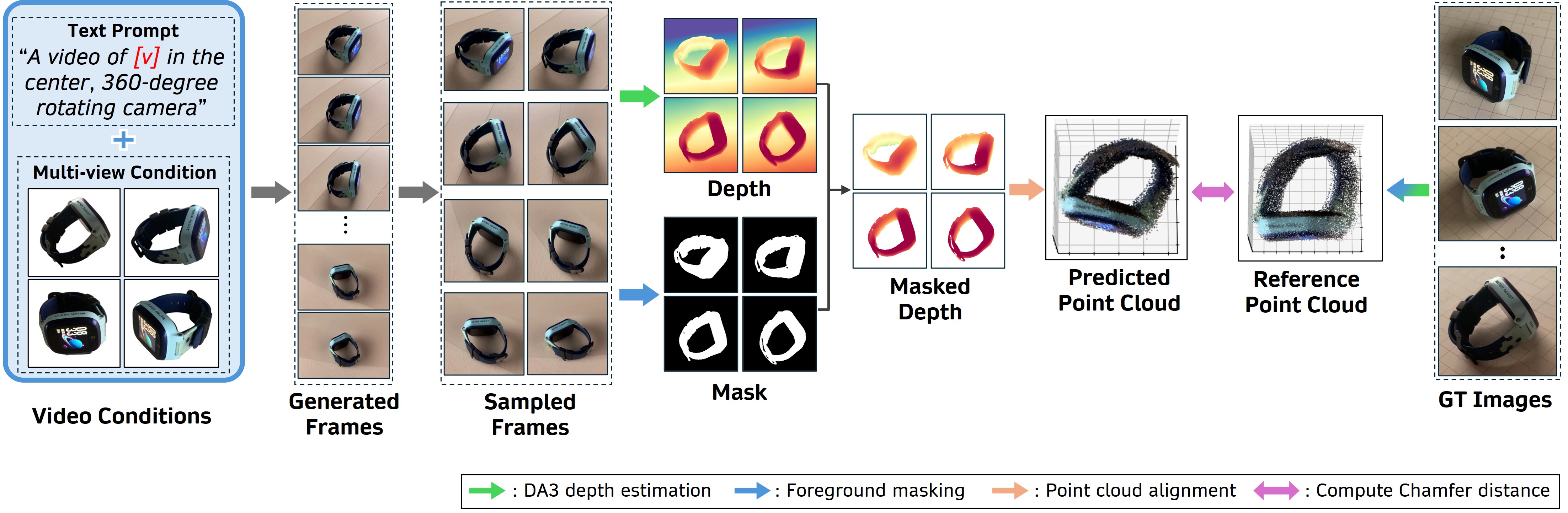}
\caption{\textbf{Pipeline of the 3D geometric fidelity evaluation.} Generated frames and ground-truth images are independently lifted to 3D point clouds using depth estimation and foreground masking. The reconstructed point clouds are then aligned to compute the Chamfer Distance.}
\label{fig:3d_geometric_consistency} 
\end{figure*}

To complement these metrics, we designed an LLM-as-a-Judge~\cite{zheng2023judging} protocol using GPT-4o~\cite{hurst2024gpt}. We
prompt GPT-4o with the four condition views alongside 10 uniformly sampled frames from the
generated video. The reference images and video frames are provided as separate images in the multimodal
message, labeled with text headers to distinguish the two groups. We use a temperature of 0.1 to reduce
output variance while allowing minimal stochasticity. Each dataset-method pair is evaluated 5 times
independently, and we report the mean and standard deviation of the scores. The model evaluates identity preservation across four dimensions:
\begin{itemize}
  \item \textbf{Shape Preservation:} Adherence to the original silhouette and overall shape across viewpoints.
  \item \textbf{Color Preservation:} Consistency of color patterns and distribution.
  \item \textbf{Detail Preservation:} Retention of fine textures, texts, and logos.
  \item \textbf{Overall Identity:} Overall recognizability of the subject.
\end{itemize}

Each dimension is scored on a 1--5 scale, where 1 indicates a completely unrecognizable subject and 5
indicates nearly perfect identity preservation. The exact GPT-4o prompt used for this evaluation is provided below:

\begin{tcblisting}{
  listing only,
  listing options={
    basicstyle=\ttfamily\small,
    breaklines=true,
    columns=fullflexible
  },
  colback=gray!5,
  colframe=gray!50,
}
You are an expert judge evaluating how well an AI-generated video 
preserves the identity of a reference subject/object.

You are given:
1. Reference images of the original subject
2. Frames extracted from a generated video
Evaluate the following aspects of Subject Identity Preservation on 
a scale of 1-5:
(1) Shape Preservation - Does the generated subject maintain the same overall shape and silhouette as the reference across viewpoints?
(2) Color Preservation - Are the colors, color patterns, and color distribution of the subject consistent with the reference?
(3) Detail Preservation - Are fine details (textures, logos, labels, patterns, small features) preserved accurately?
(4) Overall Identity - Considering all aspects holistically, would a human recognize this as the same specific object/subject?

Scoring guide:
- 1 = Completely different, unrecognizable
- 2 = Vaguely similar but major identity differences
- 3 = Recognizably similar but noticeable differences in some aspects
- 4 = Clearly the same subject with only minor differences
- 5 = Nearly perfect identity preservation

You MUST output ONLY in this exact format (no other text):
shape_preservation: <score>
color_preservation: <score>
detail_preservation: <score>
overall_identity: <score>
\end{tcblisting}

\subsection{3D Geometric Fidelity Alignment Protocol}
To explicitly evaluate the 3D structural accuracy, we designed a depth-based point cloud alignment protocol (Fig.~\ref{fig:3d_geometric_consistency}). For each object in our 
\textit{3D-CustomBench}, we obtain two sets of views: the ground-truth 
multi-view images and frames sampled from the $360^\circ$ rotation videos 
generated in Sec.~\ref{subsec:multiview_fidelity}. We independently 
process the frames using Depth Anything 3~\cite{lin2025depth} to estimate per-view metric depth maps. Concurrently, BiRefNet~\cite{zheng2024bilateral} is employed to obtain precise foreground segmentation masks. The estimated depth maps are then masked and back-projected into unified world-coordinate point clouds.

\begin{figure*}[!ht]
\centering
\includegraphics[width=\textwidth]{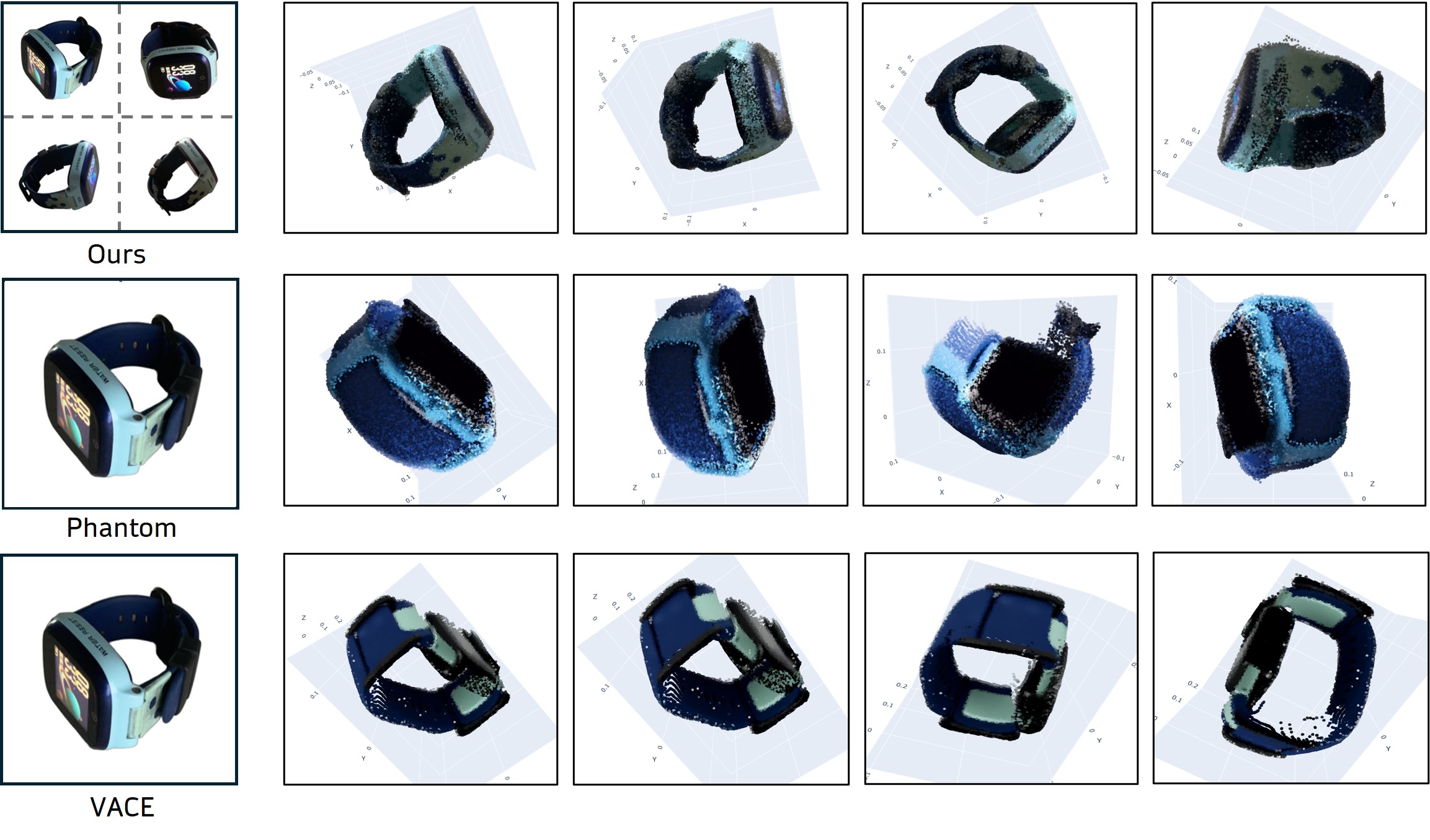}
\caption{\textbf{Qualitative comparison of reconstructed 3D point clouds.}
Using the evaluation pipeline described above, we lift each method's generated $360^\circ$ rotation video into colored point clouds and visualize them from four different viewpoints. The leftmost column shows the reference condition used for each method: for ours, a representative subset of the 30 multi-view images is displayed, while single-view baselines (Phantom~\cite{liu2025phantom} and VACE~\cite{jiang2025vace}) rely on a single reference image. Our framework produces a geometrically coherent 3D structure, while Phantom~\cite{liu2025phantom} yields over-smoothed geometry and VACE~\cite{jiang2025vace} exhibits fragmented surfaces with structural artifacts.}
\label{fig:3d_qualitative_comparison}
\end{figure*}

Since the generated point cloud ($\mathcal{P}_{gen}$) and the ground-truth point cloud ($\mathcal{P}_{gt}$) reside in different arbitrary coordinate frames, we align them via a three-stage registration process:
\begin{enumerate}
    \item \textbf{Coarse Alignment:} We extract FPFH~\cite{rusu2009fast} features and perform global registration using RANSAC~\cite{fischler1981random}.
    \item \textbf{Fine Alignment:} We apply the Iterative Closest Point (ICP)~\cite{besl1992method} algorithm for precise rigid registration.
    \item \textbf{Scale Correction:} Finally, we utilize the Umeyama similarity transformation~\cite{umeyama2002least} to correct any residual scale differences between the two spaces.
\end{enumerate}
Following this alignment, the Chamfer Distance~\cite{aanaes2016large} is computed as the average of the Accuracy and Completeness metrics to evaluate the geometric fidelity. Fig.~\ref{fig:3d_qualitative_comparison} visualizes representative point cloud reconstructions from each method, illustrating the geometric differences captured by this protocol.

\subsection{Video Quality and Text Alignment Metrics}
For each object in \textit{3D-CustomBench}, we use GPT-4o~\cite{hurst2024gpt} 
to automatically generate one validation prompt that describes a realistic, 
cinematic scenario featuring the subject in a natural context 
(e.g., human-object interactions or dynamic environments). To measure how faithfully the generated video reflects the given text prompt, we compute a video-text alignment score using ViCLIP~\cite{wang2024internvideo2}. For each generated video, we uniformly sample 8 frames and feed them jointly into the ViCLIP-L/14 vision encoder. This encoder applies temporal attention across the frames to produce a single, comprehensive video-level embedding. The final ViCLIP Score is calculated as the cosine similarity between this video-level embedding and the text embedding of the corresponding validation prompt extracted by the ViCLIP text encoder.

\end{document}